%% file: main.tex
\documentclass[10pt,twocolumn,letterpaper]{article}

\usepackage[pagenumbers]{cvpr} 

\usepackage{dsfont}
\usepackage{multirow}
\usepackage{multicol}
\usepackage{soul}
\usepackage{graphicx}
\usepackage{setspace}
\usepackage{xcolor}
\usepackage[most]{tcolorbox}
\usepackage{placeins}

\newtcolorbox{promptbox}{
  enhanced,
  breakable,
  colback=white,
  colframe=black,
  boxrule=0.4pt,
  arc=1pt,
  left=4pt,
  right=4pt,
  top=4pt,
  bottom=4pt,
}

\input{preamble}

\definecolor{cvprblue}{rgb}{0.21,0.49,0.74}
\usepackage[pagebackref,breaklinks,colorlinks,allcolors=cvprblue]{hyperref}

\def\paperID{11789} 
\def\confName{CVPR}
\def\confYear{2026}
\newcommand{\name}{ItemizedCLIP}
\newcommand{\cutabovecaption}{\vspace{-0mm}}
\newcommand{\cutbelowtable}{\vspace{-0mm}}
\newcommand{\cutabovealign}{\vspace{-0mm}}
\newcommand{\cutbelowalign}{\vspace{-0mm}}
\newcommand{\cutabovesection}{\vspace{-0mm}}
\newcommand{\cutbelowsection}{\vspace{-0mm}}
\newcommand{\cutabovesubsection}{\vspace{-0mm}}
\newcommand{\cutbelowsubsection}{\vspace{-0mm}}
\title{Learning complete and explainable visual representations from \\ itemized text supervision}

\author{Yiwei Lyu$^1$ \quad Chenhui Zhao$^1$ \quad Soumyanil Banerjee$^1$ \quad Shixuan Liu$^2$ \quad Akshay Rao$^1$ \\ Akhil Kondepudi$^1$ \quad Honglak Lee$^1$  \quad Todd C. Hollon$^1$ \\ 
\\  $^1$University of Michigan \quad $^2$University of Illinois Urbana-Champaign \\ {\tt\small yiweilyu@umich.edu}
}

\begin{document}
\maketitle

\input{0_abstract}

\input{1_main}

{
    \small
    \bibliographystyle{ieeenat_fullname}
    \bibliography{main}
}

\clearpage

\input{X_suppl}



\end{document}

%% file: preamble.tex
\newcommand{\SigLip}{\mathrm{SigL}} 



%% file: 0_abstract.tex
\begin{abstract}
Training vision models with language supervision enables general and transferable representations. However, many visual domains, especially non-object-centric domains such as medical imaging and remote sensing, contain itemized text annotations: multiple text items describing distinct and semantically independent findings within a single image. Such supervision differs from standard multi-caption supervision, where captions are redundant or highly overlapping. Here, we introduce \name, a framework for learning complete and explainable visual representations from itemized text supervision. \name~employs a cross-attention module to produce text item-conditioned visual embeddings and a set of tailored objectives that jointly enforce item independence (distinct regions for distinct items) and representation completeness (coverage of all items). Across four domains with naturally itemized text supervision (brain MRI, head CT, chest CT, remote sensing) and one additional synthetically itemized dataset, \name~achieves substantial improvements in zero-shot performance and fine-grained interpretability over baselines. The resulting \name~representations are semantically grounded, item-differentiable, complete, and visually interpretable. Our code is available at \url{https://github.com/MLNeurosurg/ItemizedCLIP}.

\end{abstract}

\vspace{-3mm}

%% file: 1_main.tex
\cutabovesection

\section{Introduction}
\label{sec:intro}

\cutbelowsection

\begin{figure}
    \centering
    \includegraphics[width=1.03\linewidth]{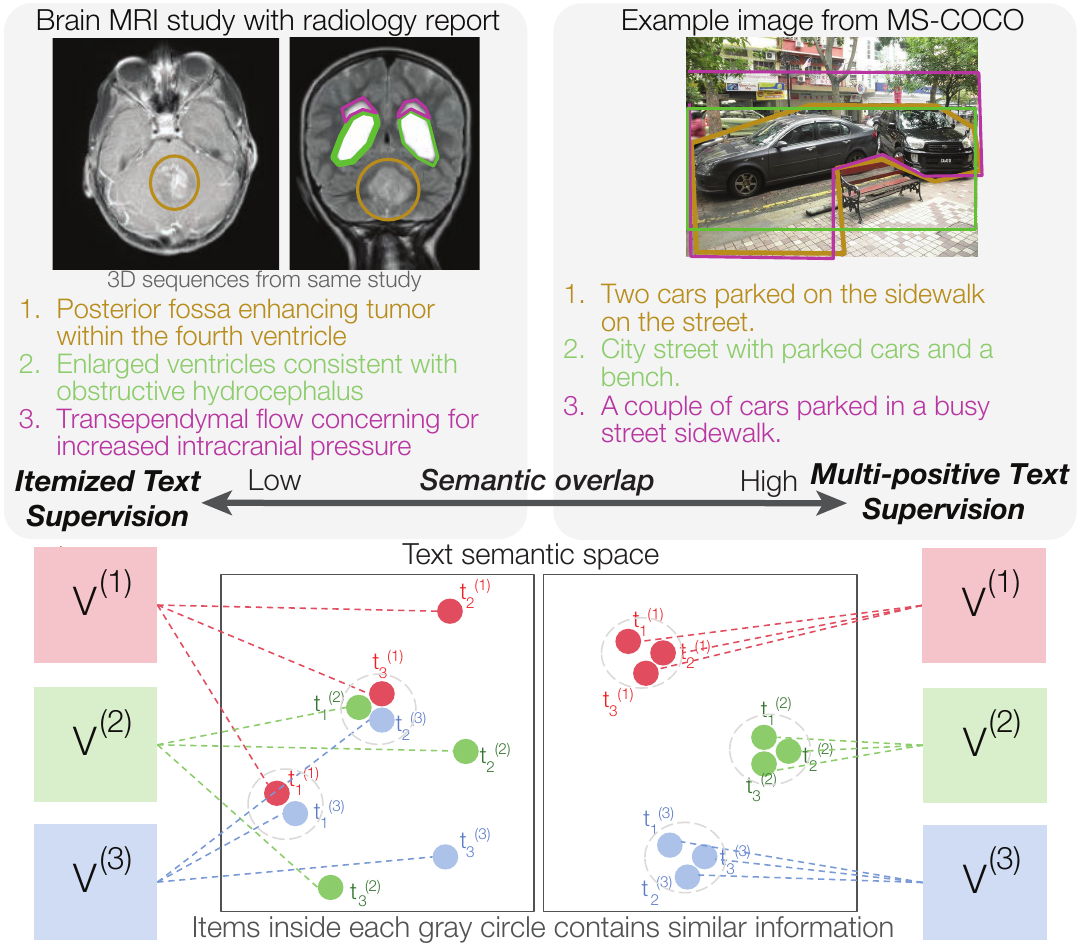}
    \cutabovecaption
    \cutabovecaption
    \cutabovecaption
    \caption{Comparison between itemized text supervision and multi-positive text supervision. (Left) In itemized supervision, each image is paired with multiple independent text items describing distinct findings or regions (e.g., different abnormalities in a brain MRI). These items have low semantic overlap and may recur across unrelated images. Effective visual representations under itemized supervision must capture information from all items rather than any single one. (Right) In multi-positive text supervision, each image is paired with several captions that describe similar visual content with high semantic overlap. (Bottom) Schematic demonstration of the relationship between itemized versus multi-positive settings in the text embedding space. Three visual examples, $V^{(i)}$, each with three associated text items/captions, $t^{(i)}_{j}$, are shown in the embedding space. Captions cluster together in multi-positive settings. Itemized text does not cluster and can share semantic information across visual examples.}
    \label{fig:intro}

    \cutbelowtable
    \cutbelowtable
\end{figure}

Language supervision has emerged as a powerful paradigm for training vision models that generalize across tasks and domains~\cite{radford2021learning}. By aligning images with textual descriptions through contrastive objectives, models such as CLIP learn representations that transfer broadly to classification, retrieval, and captioning without task-specific labels \cite{radford2021learning,zhai2023sigmoid}. Recent extensions have shown that enriching supervision with multiple positive captions per image, or multi-positive text supervision, can further improve robustness and semantic coverage~\cite{fan2023improving,lai2024veclip,zheng2024dreamlip}. In this setting, captions often express the same visual content in slightly different words, producing high semantic overlap among captions.

However, many important visual domains do not follow this paradigm. In medical imaging, remote sensing, and other non-object-centric settings, images are paired not with redundant captions but with itemized textual findings: independent statements that each describe a distinct visual scene within the same image. For instance, a brain MRI report may list separate findings such as “posterior fossa enhancing tumor,” “ventricular enlargement,” and “transependymal flow.” These text items have minimal semantic overlap and refer to spatially disjoint regions. We call this setting \textbf{itemized text supervision} (Figure~\ref{fig:intro}).

Itemized text supervision presents a unique learning challenge. Under multi-positive text supervision, a model is rewarded for matching an image with any one of its captions, since they all describe similar content. In contrast, under itemized supervision, a model must represent all text items simultaneously: omitting any one may correspond to missing a critical finding, such as a tumor or hemorrhage in a brain MRI, leading to a misdiagnosis. Effective representations must therefore satisfy two key properties:

\begin{itemize}
    \item \textit{Item independence}: features associated with different text items should remain distinguishable and localized to their corresponding regions.
    \item \textit{Representation completeness}: the combined visual embedding should encode information from all items, ensuring full coverage of visual content.
\end{itemize}

Existing CLIP-style and multi-positive approaches do not satisfy these criteria. For example, multi-positive contrastive objectives explicitly pull together embeddings of all positive captions for the same image, contradicting the independence requirement. Prior models trained for radiology using CLIP-style objectives concatenate all text items into a single caption ~\cite{lyu2025learning,zhao2025scalable, hamamci2024developing}, discarding the compositional structure inherent in itemized supervision. 

To address this gap, we propose \name, a framework for learning complete and explainable visual representations directly from itemized text supervision. \name~employs a cross-attention module to generate item text-conditioned visual embeddings and introduces a set of training objectives specifically designed for this regime. These include an Itemized Local Alignment (ILA) objective that aligns each text item with its corresponding visual evidence with additional mechanisms for completeness and robustness, an Inter-Item Separation (IIS) loss that enforces differentiation between items, a Multi-Positive Siglip (MPS) objective for global alignment, and a Key Token Alignment (KTA) objective to reinforce alignment between each text item and the corresponding small subset of visual tokens. Together, these components allow \name~to learn localized, granular, comprehensive, and interpretable visual features without explicit region annotations.

We evaluate \name~across four domains with naturally itemized text supervision (brain MRI, head CT, chest CT, and remote sensing) and one additional dataset with synthetically itemized captions. \name~outperforms CLIP-style models and other baselines in zero-shot evaluations across all domains, visualization analyses reveal that \name’s attention maps align with human-interpretable findings, demonstrating inherent explainability and grounding. Our main contributions are:

\begin{enumerate}
    \item We formalize itemized text supervision as a distinct and practically important paradigm for language-supervised learning, and this underexplored paradigm naturally occurs in many application domains.
    \item We introduce \name, a method that jointly enforces item independence and representation completeness through cross-attention and new loss formulations.
    \item We provide comprehensive empirical and qualitative evaluations across five domains with itemized text supervision, and \name~consistently achieves strong zero-shot performance, item differentiability, representation completeness and interpretability.
\end{enumerate}

\begin{table}[]
\scalebox{0.85}{
\begin{tabular}{@{\hspace{0pt}}l@{\hspace{1pt}}|@{\hspace{1pt}}c@{\hspace{3pt}}c@{\hspace{3pt}}c@{\hspace{3pt}}c@{\hspace{0pt}}}
\toprule
 & Llip & Dream- & FLAIR & Itemiz- \\
 & \cite{lavoie2024modeling}& LIP~\cite{zheng2024dreamlip} & \cite{xiao2025flair} & edCLIP \\ \hline
Text-conditioned visual representations & $\textcolor{green}{\checkmark}$ & $\textcolor{green}{\checkmark}$ & $\textcolor{green}{\checkmark}$ & $\textcolor{green}{\checkmark}$ \\
Visual token localization from text & $\textcolor{red}\times$ & $\textcolor{green}{\checkmark}$ & $\textcolor{green}{\checkmark}$ & $\textcolor{green}{\checkmark}$ \\
Zero-shot with local visual tokens & $\textcolor{red}\times$ & $\textcolor{red}\times$ & $\textcolor{green}{\checkmark}$ & $\textcolor{green}{\checkmark}$ \\
Enforces item independence & $\textcolor{red}\times$ & $\textcolor{green}{\checkmark}$ & $\textcolor{red}\times$ & $\textcolor{green}{\checkmark}$ \\
Enforces representation completeness & $\textcolor{red}\times$ & $\textcolor{red}\times$ & $\textcolor{red}\times$ & $\textcolor{green}{\checkmark}$ \\ \bottomrule
\end{tabular}
}
\cutabovecaption
\caption{\name~is designed specifically for itemized text supervision, which differs from multi-positive supervision assumed in previous works, and enforces both item independence and representation completeness.}
\label{tab:rel}
\cutbelowtable
\cutbelowtable
\end{table}

\cutabovesection

\section{Related Work}

\cutbelowsection

We provide a comprehensive discussion of related work in Appendix~\ref{app:rel}. Here we discuss a few existing works designed for multi-positive supervision that are most closely related to \name: Llip~\cite{lavoie2024modeling} first proposed to address diversity in multiple positive captions via text-conditioned visual representations generated by a cross-attention with text encoding as key and generated visual mixture tokens as key and value. DreamLIP~\cite{zheng2024dreamlip} proposed a text-conditioned visual representation based on cross attention between subcaption encodings and local visual tokens (instead of a few mixture tokens in Llip) to allow localization of subcaptions to specific local tokens. FLAIR~\cite{xiao2025flair} modifies the training objective from DreamLIP with different negative pairs (TCS loss) and combines it with Multi-positive SigLIP. These works have made significant progress in learning explainable and localized visual representations from multiple positive long captions, but are not designed for itemized text supervision. \name~adapts FLAIR's TCS+MPS objective and added additional objectives to learn visual representations from itemized text supervision that are not only high-quality, explainable and localizable, but also satisfy the requirements of itemized text supervision (item differentiability and representation completeness). We summarize the comparison between these works in Table~\ref{tab:rel}.

\begin{figure*}
    \centering
    \includegraphics[width=1.0\linewidth]{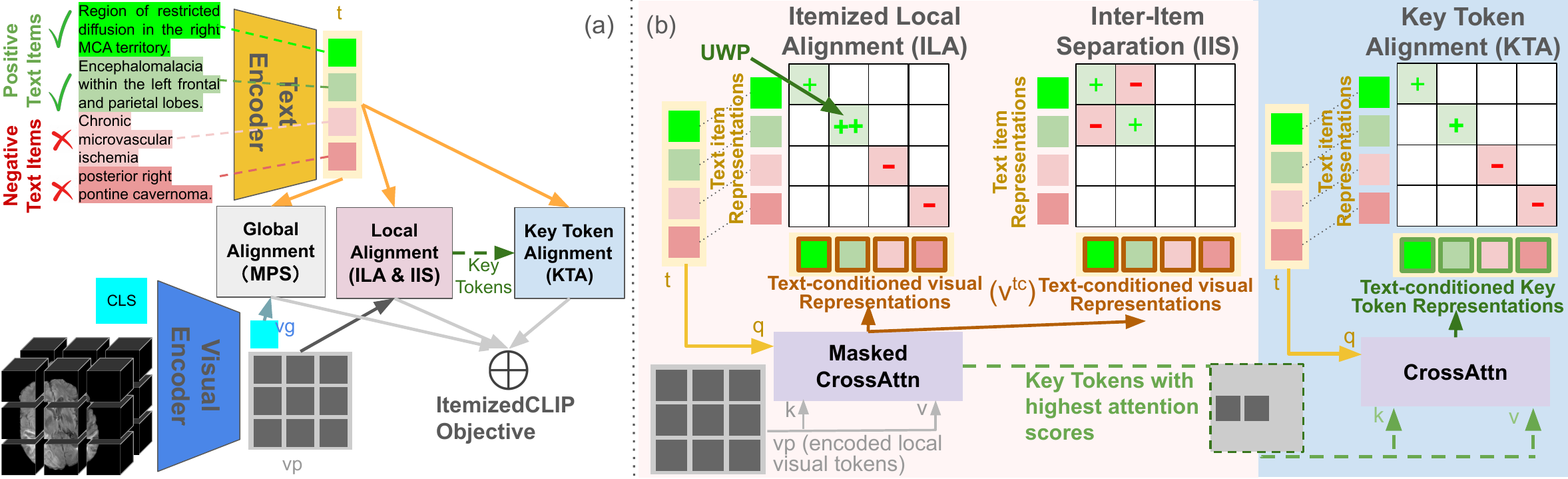}
    \cutabovecaption
    \cutabovecaption
    \cutabovecaption
    \caption{Overview of \name. \textbf{(a)} \name~includes several SigLIP-based objective components: Itemized Local Alignment \textbf{(ILA)}, Inter-Item Separation (\textbf{IIS}), Multi-positive SigLIP \textbf{(MPS)} and Key Token Alignment \textbf{(KTA)}, and the overall objective is their weighted sum. MPS follows~\cite{xiao2025flair} and applies a SigLIP objective between the global visual token and each text item representation. \textbf{(b)} ILA, IIS and KTA losses are all based on cosine similarities between text item encodings and cross-attention output between text encodings and encoded visual tokens. The grids show which cosine similarities are being used as positive (\textbf{\textcolor{green}{+}}) / negative (\textbf{\textcolor{red}{-}}) pairs for each objective. ILA mainly aims to distinguish positive text items from negative ones, while IIS aims to distinguish between positive text items. Upweighting Worst Positive \textbf{(UWP)} upweights (\textbf{\textcolor{green}{++}}) the current weakest positive pair in ILA when computing the ILA loss. The key visual tokens from ILA's masked cross attention with the highest attention scores are used for KTA, which computes an objective similar to ILA but with key tokens only and no masking.}
    \label{fig:methods}
    \cutbelowtable
    \cutbelowtable
\end{figure*}

\cutabovesection

\section{\name~Methodology}

\cutbelowsection

\cutabovesubsection

\subsection{Architecture Overview}

\cutbelowsubsection

Like most CLIP-based frameworks, \name~consists of a visual encoder $E_V$ (vision transformer with CLS token) and a text encoder $E_T$. \name~also contains an additional multi-headed attention $\texttt{CrossAttn}$ with linear qkv projection layers. The training data is a paired visual-language  $D=\{V^{(i)},T^{(i)}\}^{N}_{i=1}$ under itemized text supervision, where each visual input $V^{(i)}$ can be patched into $m$ patches $V^{(i)}=\{p^{(i)}_1,p^{(i)}_{2},...,p^{(i)}_m\}$ and each text input $T^{(i)}$ contains $n_i$ text items $T_i=\{item^{(i)}_{1},item^{(i)}_2,...,item^{(i)}_{n_i}\}$. $n_i$ may vary across different $T^{(i)}$.

During the forward pass, given $(V^{(i)},T^{(i)})$, we encode the text representation $t^{(i)}_j$ for each $item^{(i)}_j$ separately: $t^{(i)}_j=E_T(item^{(i)}_j)$, and we obtain two visual representations $(vg^{(i)},vp^{(i)}) = E_V(V^{(i)})$ where $vg^{(i)}$ is the global visual representation obtained from the CLS token output of $E_V$, while $vp^{(i)}=\{vp^{(i)}_1,vp^{(i)}_2,...,vp^{(i)}_m\}$ is the patch-level (i.e. visual-token level) visual representation from the final transformer layer output of $E_V$ over each of the $m$ patches.

\cutabovesubsection

\subsection{\name~Training Objective}

\cutbelowsubsection

\name~introduces a family of objectives tailored to the characteristics of itemized text supervision (Figure~\ref{fig:methods}). 


\textbf{(a) Itemized Local Alignment (ILA)} We begin from Text Conditioned SigLIP (TCS) from FLAIR~\cite{xiao2025flair} and adapted it for itemized text supervision. TCS first generates a text-conditioned image representation and computes its cosine similarity with the text item: given text item $t$ and visual patch representations $vp$, we compute $TCSim(t,vp) = CS(t,\texttt{CrossAttn}(t,vp,vp))$ (where $CS$ is cosine similarity, and \texttt{CrossAttn} is the multi-head cross-attention mechanism that uses text item representation $t$ to query visual patch representations $vp$ to compute the text-conditioned visual representation). TCS uses a SigLIP objective to maximize $TCSim(t,vp)$ if the text item belongs to the visual, and minimize $TCSim(t,vp)$ otherwise. In practice, when given an input batch $B=\{V^{(i)},T^{(i)}\}^{|B|}_{i=1}$, for each i, we compute $TCSim(t^{(i)}_j,vp^{(i)})$ for all $j=1,2,...,n_i$ as positive pairs to be maximized, and we randomly sample one text item representation from each other visual-text pair in the batch ($t^{(k)}_r$, where $k\not=i$ and $r \sim \texttt{random}(1,2,...,n_k)$) as negative pairs with $vp^{(i)}$ (i.e. minimizing $TCSim(t^{(k)}_r,vp^{(i)})$) for computational efficiency. So the overall TCS objective for the batch would be:
{
\small
\begin{align}
\mathcal{L}_{\text{TCS}}(B)
&= -\frac{1}{|B|}\sum_{i=1}^{|B|} \Bigg[
      \sum_{j=1}^{n_i} \SigLip \big(TCSim(t^{(i)}_j, vp^{(i)}), 1,b,\tau\big) \notag\\
&\quad + \sum_{k \neq i} \SigLip\big(TCSim(t^{(k)}_{r_k}, vp^{(i)}), -1,b,\tau\big)
   \Bigg] \nonumber
\end{align}}

\noindent where $r_k$ is a random integer between 1 and $n_i$, $\SigLip$ represents the single-pair SigLIP objective $\SigLip(k,z,b,\tau)=\log(\frac{1}{1+e^{z(-\tau k+b)}})$, $b$ and $\tau$ are trainable bias and temperature parameters, and $z$ is the sign variable.

In \name, we make two modifications to TCS to obtain the ILA objective. To further encourage the model to generate visual representations that captures information from all positive text items (\emph{completeness} criteria), we perform \textbf{Upweighting Worst Positive (UWP)}: for each visual $V^{(i)}$, we upweight the positive pair loss between $vp^{(i)}$ and its corresponding text item with the lowest $TCSim$ by a factor of $w_{\text{uwp}}$, which is a hyperparameter. In addition, we perform masked attention within $TCSim$ calculation during training with a randomly generated binary mask from $\text{Bernoulli}(p_{\text{mask}})$, where $p_{\text{mask}}$ is a hyperparameter controlling the level of masking. So ILA (i.e. TCS+UWP+masking) is:



\cutabovealign
{
\footnotesize
\begin{align}
&\mathcal{L}_{\text{ILA}}(B) = \nonumber -\frac{1}{|B|}\sum_{i=1}^{|B|} \Bigg[
      \sum_{j=1}^{n_i} w^{(i)}_j \SigLip\big(TCSim_{\text{masked}}(t^{(i)}_j, vp^{(i)}, \notag\\
&\quad p_{\text{mask}}),1,b,\tau\big) + \sum_{k \neq i}  \SigLip\big(TCSim_{\text{masked}}(t^{(k)}_{r_k}, vp^{(i)},p_{\text{mask}}), -1,b,\tau\big)
   \Bigg] \nonumber
\end{align}
}
\cutbelowalign

\noindent where $w^{(i)}_j = \begin{cases}  w_{\text{uwp}} & j = \text{argmin}_j TCSim(t^{(i)}_j,vp^{(i)}) \\ 1 & \text{otherwise} \end{cases}$.

UWP forces the model to learn more from the positive pairs that it currently thinks is the ``least positive", therefore improving representation completeness. The masked attention prevents each attention head within the $\texttt{CrossAttn}$ module from seeing a random subset of the visual tokens during training, thereby preventing overfitting and yielding more robust visual grounding.

\textbf{(b) Inter-Item Separation (IIS)} The item-independence property implies that different text items should attend to different regions of the image, so we introduce an additional inter-item loss to encourage this. IIS works by treating the text-conditioned visual representation $v^{tc}_{i,j}=\texttt{CrossAttn}(t^{(i)}_j,vp^{(i)},vp^{(i)})$ and a text item from the same visual $t_k^{(i)}$ as negative pairs if $k\not=j$ and as positive pairs if $k=j$. Formally, we have

\cutabovealign
{\small 
\begin{align}
\mathcal{L}_{\text{IIS}}(B) &= -\frac{1}{|B|}\sum_{i=1}^{|B|}\sum_{j=1}^{n_i}\sum_{k=1}^{n_i} \SigLip(CS(t^{(i)}_k,v^{tc}_{i,j}),\mathcal{I}_{k=j},b,\tau)  \nonumber
   \end{align}
}


where $\mathcal{I}_{k=j}=1$ if $k=j$ and is -1 otherwise. During training, we reuse the masked cross-attention results from ILA. IIS loss enhances text item differentiability by effectively pushing apart $v^{tc}_{i,j}$ and $v^{tc}_{i,k}$ for all $j\not=k$, thereby forcing the cross-attention module to attend to distinct parts of the visual representation for different text items. FLAIR~\cite{xiao2025flair} explored an IIS-like objective in its Appendix but abandoned it as it didn't work well under multi-positive supervision. However, we found that the IIS objective is very effective for itemized text supervision (see section~\ref{sec:differentiability}).

\textbf{(c) Multi-positive SigLIP (MPS)} Multi-positive SigLIP loss is the SigLIP loss defined under multi-positive supervision, where the global visual representation $vg^{(i)}$: 

{\small 
\begin{align}
\mathcal{L}_{\text{MPS}}(B) &= -\frac{1}{|B|}\sum_{i=1}^{|B|}\Bigg[\sum_{j=1}^{n_i} \SigLip(CS(t^{(i)}_j,vg^{(i)}),1,b,\tau) \nonumber\\
&+\sum_{k \neq i} \SigLip\big(CS(t^{(k)}_{r_k}, vg^{(i)}), -1,b,\tau\big)
   \Bigg] \nonumber
   \end{align}
}

We follow~\cite{xiao2025flair} and only include one random text item from every other visual in the batch as negative pairs. Although MPS has the effect of pushing representations of different text items paired to the same visual (i.e. $t^{(i)}_j$ with different $j$) closer to each other, which is undesirable in itemized text supervision due to item independence, we empirically found that adding lower-weighted MPS loss help improve the model's awareness of global visual properties which benefits downstream zero-shot classifications.

\textbf{(d) Key Token Alignment (KTA)} KTA loss guides the model to align text items with compact, high-attention visual tokens. For each text item representation $t$ and visual tokens $vp$, we obtain the attention map from $\texttt{CrossAttn(t,vp,vp)}$ (mean-aggregated across all attention heads) and define $KT(t,vp)$ to be the subset of $vp$ that has the top $K\%$ attention scores in the attention map, where $K$ is a hyperparameter. To strengthen localization, we perform TCS loss with only the key tokens ($KT(t,vp)$ instead of $vp$):

\cutabovealign
{
\footnotesize
\begin{align}
\mathcal{L}_{\text{KTA}}(B)
&= -\frac{1}{|B|}\sum_{i=1}^{|B|} \Bigg[
      \sum_{j=1}^{n_i} \SigLip\big(TCSim(t^{(i)}_j, KT(t^{(i)}_j,vp^{(i))}),1,  \notag\\
\quad & b,\tau\big) + \sum_{k \neq i} \SigLip\big(TCSim(t^{(k)}_{r_k}, KT(t^{(k)}_{r_k},vp^{(i))}), -1,b,\tau\big)
   \Bigg] \nonumber
\end{align}
}

\textbf{(e) Overall loss} of \name~is the weighted sum of the four objectives ((a)-(d)):
{
\begin{align}
\mathcal{L}_{\text{all}}=\mathcal{L}_{\text{ILA}}+\lambda_{\text{IIS}}\mathcal{L}_{\text{IIS}}+\lambda_{\text{MPS}}\mathcal{L}_{\text{MPS}}+\lambda_{\text{KTA}}\mathcal{L}_{\text{KTA}} \nonumber  \end{align}}
\cutabovealign
\cutabovesubsection

\subsection{Zero-shot inference}

\cutbelowsubsection

When performing zero-shot inference with \name, given a visual $V^{(i)}$ and a set of classes in text description $c_1,c_2,...$, we first encode them with the visual/text encoders to obtain visual tokens $vp^{(i)}$ and text representations $t_{c_1},t_{c_2}, ...$, and then our prediction logit for class $k$ is simply $TCSim(t_{c_k},vp^{(i)})$.

\begin{table*}[]
\centering
\scalebox{0.85}{
\begin{tabular}{l|ccccccccccc|c}
\toprule
 Categories & Cyst & \begin{tabular}[c]{@{}c@{}}Develo-\\ pmental\end{tabular} & \begin{tabular}[c]{@{}c@{}}Infec-\\ tious\end{tabular} & \begin{tabular}[c]{@{}c@{}}Inflam-\\ matory\end{tabular} & Sellar & Spine & Structural & Surgical & Trauma & Tumor & Vascular & \begin{tabular}[c]{@{}c@{}}Overall \\ mAUC\end{tabular} \\ \hline
 \# tasks in category & 2 & 3 & 2 & 2 & 3 & 1 & 8 & 3 & 3 & 13 & 12 & 52 \\ \hline
Prima~\cite{lyu2025learning} & 64.4 & 83.2 & 69.7 & 81.6 & 79.9 & 85.7 & 68.7 & 75.2 & 69.0 & 80.3 & 75.4 & 75.7 \\
FullViT~\cite{zhao2025scalable} & 76.1	&\underline{89.6}	&76.3	& \underline{93.4}	&\underline{88.7}	&81.7& 	76.8&  	\underline{88.0}	&  75.4  &	88.3	&\underline{78.5}&	82.7                \\
HLIP~\cite{zhao2025scalable} & \underline{77.8} & 87.2 & \underline{77.2} & \textbf{91.2} & 88.6 & \underline{86.1} & \underline{80.7} & 85.6 & \underline{81.3} & \underline{89.7} & 77.6 & \underline{83.7} \\
\name~(Ours) & \textbf{84.6} & \textbf{96.8} & \textbf{88.4} & 87.0 & \textbf{91.2} & \textbf{94.3} & \textbf{88.7} & \textbf{95.3} & \textbf{85.9} & \textbf{92.5} & \textbf{89.1} & \textbf{90.5} \\ \bottomrule
\end{tabular}
}
\cutabovecaption
\caption{ Results for zero-shot evaluation on UM220K prospective test set~\cite{lyu2025learning}. The numbers reported are mean AUC across tasks within each of the 11 categories, as well as across all 52 tasks. Individual task AUC is reported in Figure~\ref{fig:radar-mri} in Appendix.} 
\label{tab:mri-prospective}
\cutbelowtable
\end{table*}

\begin{table*}[]
\centering
\scalebox{0.85}{
\begin{tabular}{l|cccccc|cccccc|c}
\toprule
  & \multicolumn{6}{c|}{Pub-Brain-5-gt} & \multicolumn{6}{c|}{Pub-Brain-5} & \multicolumn{1}{c}{\multirow{2}{*}{\begin{tabular}[c]{@{}c@{}}12-metric \\ Mean\\ BAcc\end{tabular}}} \\
 & \multicolumn{1}{c}{\begin{tabular}[c]{@{}c@{}}Str-\\ oke\end{tabular}} & \multicolumn{1}{c}{\begin{tabular}[c]{@{}c@{}}Gli-\\ oma\end{tabular}} & \multicolumn{1}{c}{\begin{tabular}[c]{@{}c@{}}Menin-\\ gioma\end{tabular}} & \multicolumn{1}{c}{\begin{tabular}[c]{@{}c@{}}Meta-\\ stasis\end{tabular}} & \multicolumn{1}{c}{\begin{tabular}[c]{@{}c@{}}Tumor \\ (3-way)\end{tabular}} & \multicolumn{1}{c|}{\begin{tabular}[c]{@{}c@{}}Disease \\ (5-way)\end{tabular}} & \multicolumn{1}{c}{\begin{tabular}[c]{@{}c@{}}Str-\\ oke\end{tabular}} & \multicolumn{1}{c}{\begin{tabular}[c]{@{}c@{}}Gli-\\ oma\end{tabular}} & \multicolumn{1}{c}{\begin{tabular}[c]{@{}c@{}}Menin-\\ gioma\end{tabular}} & \multicolumn{1}{c}{\begin{tabular}[c]{@{}c@{}}Meta-\\ stasis\end{tabular}} & \multicolumn{1}{c}{\begin{tabular}[c]{@{}c@{}}Tumor \\ (3-way)\end{tabular}} & \multicolumn{1}{c|}{\begin{tabular}[c]{@{}c@{}}Disease \\ (5-way)\end{tabular}} & \multicolumn{1}{c}{} \\ \hline

BiomedCLIP~\cite{zhang2023biomedclip} & 66.7 & 88.2 & 63.6 & 59.8 & 50.4 & 31.5 & 64.7 & 87.8 & 63.6 & 59.8 & 50.4 & 31.5 & 59.8 \\
 \quad+annotation & 86.4 & 94.1 & 75.8 & 75.0 & 45.7 & 45.3 & - & - & - & - & - & - & - \\
ConceptCLIP~\cite{nie2025conceptclip} & 69.6 & 92.1 & 57.8 & 69.5 & 35.2 & 31.6 & 66.8 & 91.9 & 57.7 & 67.9 & 35.7 & 30.9 & 58.9 \\
 \quad+annotation & 93.6 & \textbf{97.8} & 70.8 & \underline{76.8} & 39.4 & 50.8 & - & - & - & - & - & - & - \\
Prima~\cite{lyu2025learning} & 61.2 & 81.0 & \underline{87.7} & 53.4 & 45.9 & 31.4 & 78.8 & 89.3 & 70.8 & 64.7 & 42.8 & 31.6 & 61.6 \\
FullViT~\cite{zhao2025scalable} & 76.7 & 93.5 & 58.2 & 58.2 & 42.1 & 43.5 & 72.8 & \underline{93.4} & 72.9 & 63.1 & 45.7 & 43.4 & 63.6 \\
HLIP~\cite{zhao2025scalable} & \underline{95.0} & 89.2 & 79.6 & 73.4 & \underline{54.8} & \underline{61.3} & \underline{91.5} & 89.2 & \underline{79.2} & \underline{78.1} & \textbf{63.3} & \underline{63.9} & \underline{76.5} \\
\name~(Ours) & \textbf{99.0} & \underline{96.9} & \textbf{89.3} & \textbf{94.8} & \textbf{58.2} & \textbf{68.2} & \textbf{97.3} & \textbf{96.8} & \textbf{89.4} & \textbf{88.8} & \underline{57.2} & \textbf{67.6} & \textbf{83.6}  \\ \bottomrule
\end{tabular}
}\cutabovecaption 
\caption{Results for zero-shot evaluation on Pub-Brain-5 and its subset Pub-Brain-5-gt with 2D slice annotations. We follow~\cite{zhao2025scalable} to report balanced accuracy on each of the 4 diseases, as well as 3-way tumor classification and 5-way disease classification (normal + 4 diseases).
All numbers except~\name~ was reported from~\cite{zhao2025scalable}. The 2D slice annotations were only used for selecting the relevant 2D slices for BiomedCLIP and ConceptCLIP zero-shot (in the ``+annotation" rows), and are not used for any other rows.
}
\label{tab:pub-brain-5}
\cutbelowtable
\cutbelowtable
\end{table*}

\cutabovesection

\section{Experiments}

\cutbelowsection

We comprehensively evaluate \name~against SOTA baselines in 4 domains with naturally itemized text supervision: whole-study brain MRI, whole-study head CT, chest CT, and remote sensing. In addition, we create a dataset with synthetic itemized text supervision for natural images, and evaluate \name~against the same baselines.

\cutabovesubsection

\subsection{Whole-study Brain MRI}

\cutbelowsubsection

Brain MRI is one of the most important medical modalities for diagnosing brain diseases. A brain MRI contains multiple 3D volumes, or sequences, as part of a single medical imaging study. UM220K~\cite{lyu2025learning} is the largest dataset of whole-study brain MRI with paired itemized radiology reports, with over 220K study-report pairs (including 3.62M 3D sequences). We train \name~on UM220K with similar architecture and preprocessing as HLIP~\cite{zhao2025scalable}. See Appendix~\ref{app:mri-implementation} for full implementation and baseline details. Then, we perform zero-shot evaluation on UM220K prospective test set~\cite{lyu2025learning} (30K temporally separated whole-studies acquired after the training data, with annotations for 52 diagnostic tasks) and Pub-Brain-5~\cite{zhao2025scalable} (we follow~\cite{zhao2025scalable} and report performance on 12 metrics across Pub-Brain-5 and its slice-annotated subset, Pub-Brain-5-gt) and we compare \name's performance to current SOTA whole-study brain MRI models like Prima~\cite{lyu2025learning} and HLIP~\cite{zhao2025scalable}, which are also trained on UM220K. Our zero-shot prompts are identical to Prima and HLIP for fairness of comparison.

We show the results for prospective test set in Table~\ref{tab:mri-prospective} and for Pub-Brain-5 in Table~\ref{tab:pub-brain-5} against SOTA baselines. 
\name~outperformed baselines by large margins overall: +6.8\% zero-shot mean AUC across 52 tasks on the prospective test set and +7.1\% zero-shot mean balanced accuracy on Pub-Brain-5 compared to previous SOTA, and up to +18\% gains on individual tasks.


\cutabovesubsection

\subsection{Whole-study Head CT}

\cutbelowsubsection

Head CTs also contain several 3D CT sequences in each study. Each 3D CT sequence is presented in 3 different windowing levels: brain, blood, and bone. One of the largest whole-study head CT datasets with paired itemized radiology reports is HeadCT240K~\cite{zhao2025scalable}, which contains over 240K study-report pairs. We use \name~to train a model on this dataset and evaluate on downstream classification tasks against baselines. Our implementation of \name~was identical to HLIP in preprocessing and visual backbone design. We include full implementation details of \name~ in Appendix~\ref{app:ct-implementation}. We perform zero-shot evaluation on the following downstream datasets: (1) prospective test set of HeadCT240K~\cite{zhao2025scalable} with $>$21K head CT studies with annotations on 83 different diagnostic tasks; (2) RSNA~\cite{flanders2020construction}, with $>$10K head CT studies ($>$25K sequences) with annotations for 5 diagnostic tasks about intracranial hemorrhage; and (3) CQ500~\cite{chilamkurthy2018deep}, with $\sim$500 head CT studies with annotations on 10 diagnostic tasks. 

We present the prospective test set results in Table~\ref{tab:ct-prospective} and RSNA/CQ500 results in Table~\ref{tab:rsnacq500} against SOTA baselines. ItemizedCLIP again outperformed zero-shot baselines by a large margin: +9.3\% in mean AUC across 83 tasks in the prospective test set and over 5\% gains on average over all RSNA/CQ500 tasks, compared to HLIP. RSNA and CQ500 are widely used public benchmarks for evaluating head CT models, so we also compared our performance to recent head CT foundation models, including FM-HeadCT~\cite{zhu20253d}, Google-CT~\cite{yang2024advancing} and Merlin~\cite{blankemeier2024merlin}. Only linear probing performance was available from these foundation models, but \name~was able to outperform their linear probing performances with zero-shot.

\begin{table*}[]
\scalebox{0.85}{
\begin{tabular}{l|ccccccccccc|c}
\toprule
 & Cystic & Vascular & Trauma & \begin{tabular}[c]{@{}c@{}}Struc-\\ tural\end{tabular} & Surgical & Tumor & \begin{tabular}[c]{@{}c@{}}Degen-\\ erative\end{tabular} & ENT & \begin{tabular}[c]{@{}c@{}}Infec-\\ tious\end{tabular} & \begin{tabular}[c]{@{}c@{}}Cong-\\ enital\end{tabular} & Orbital & \begin{tabular}[c]{@{}c@{}}Overall\\ mAUC\end{tabular} \\ \hline
\# tasks in category & 4 & 18 & 11 & 7 & 11 & 10 & 2 & 12 & 3 & 2 & 3 & 83 \\ \hline
FullViT & \underline{68.0}	& 76.7& 	 75.3& 	75.5&  	\underline{75.5}	&  67.7	&  69.2& 	69.5& 	\underline{73.5}& 	79.6	& 64.3	&73.0 \\
HLIP & 67.5 & \underline{77.0} & \underline{78.0} & \underline{78.7} & 75.3 & \underline{73.8} & \underline{71.3} & \underline{74.6} & 72.3 & \textbf{81.4} & \underline{80.3} & \underline{75.8} \\
\name~(Ours) & \textbf{71.8} & \textbf{85.7} & \textbf{87.2} & \textbf{87.1} & \textbf{89.7} & \textbf{78.4} & \textbf{81.1} & \textbf{86.8} & \textbf{84.5} & \underline{80.1} & \textbf{91.8} & \textbf{85.1} \\ \bottomrule
\end{tabular}
}
\cutabovecaption 
\caption{Results for zero-shot evaluation HeadCT240K's prospective test set~\cite{zhao2025scalable}. The numbers reported are mean AUC across tasks within each of the 11 categories, as well as across all 83 tasks. Individual task AUC is reported in Figure~\ref{fig:radar-ct} in Appendix.} 
\label{tab:ct-prospective}
\cutbelowtable
\cutbelowtable
\cutbelowtable

\end{table*}

\begin{table}[]
\centering
\scalebox{0.85}{
\begin{tabular}{ll|cc}
\toprule
\multirow{3}{*}{Models} & \multirow{3}{*}{Inference Type} & \multicolumn{1}{l}{RSNA} & \multicolumn{1}{l}{CQ500} \\
 &  & \multicolumn{1}{l}{\begin{tabular}[c]{@{}l@{}}5-tasks \\ mAUC\end{tabular}} & \multicolumn{1}{l}{\begin{tabular}[c]{@{}l@{}}10-tasks \\ mAUC\end{tabular}} \\ \hline
FullViT~\cite{lyu2025learning} & Zero Shot & 83.3 & 73.1 \\
HLIP~\cite{lyu2025learning} & Zero Shot & 85.7 & 83.1 \\
\name~(Ours) & Zero Shot & \textbf{91.5} & \textbf{90.0} \\ \hline
FM-HeadCT~\cite{zhu20253d} & Linear Probing & 91.3 & 80.0 \\
Google-CT~\cite{yang2024advancing} & Linear Probing & 87.2 & 76.1 \\
Merlin~\cite{blankemeier2024merlin} & Linear Probing & 73.6 & 55.1 \\ \bottomrule
\end{tabular}
}\cutabovecaption 
\caption{Results for zero-shot evaluation on head CT public benchmarks RSNA~\cite{flanders2020construction} and CQ500~\cite{chilamkurthy2018deep}. We report mean AUC across all tasks within each dataset. Per-task results are included in Table~\ref{tab:rsnafull} and Table~\ref{tab:cq500full} in Appendix~\ref{app:rsnacq500}. We also include \textbf{linear probing} performance from three recent head CT foundation models, which were outperformed by \name~\textbf{zero-shot}.}
\label{tab:rsnacq500}
\cutbelowtable
\end{table}

\cutabovesubsection

\subsection{Single-sequence Chest CT}

\cutbelowsubsection

Another domain where medical images are paired with itemized text descriptions is Chest CT. We follow~\cite{zhao2025scalable} to train \name~on CT-Rate~\cite{hamamci2024developing} dataset, which contains over 25K single-sequence Chest CT scans with paired radiologist reports. Each report has an LLM-summarized version that is itemized. We implement \name~using similar data preprocessing steps and visual backbone architecture from current SOTA, HLIP~\cite{zhao2025scalable} (see Appendix~\ref{app:chestct-implementation} for implementation details), and we perform the same evaluations as~\cite{zhao2025scalable}: zero-shot classification on test split of CT-Rate~\cite{hamamci2024developing} with 16 diagnostic tasks as well as an external dataset, Rad-ChestCT~\cite{draelos2021machine}, with 14 diagnostic tasks. The results are shown in Table~\ref{tab:chestct}. \name~again outperforms all baselines across all metrics.

\begin{table}[]
\centering
\scalebox{0.85}{
\begin{tabular}{@{\hspace{0pt}}l|c@{\hspace{7pt}}c@{\hspace{7pt}}c|ccc@{\hspace{0pt}}}
\toprule
 & \multicolumn{3}{c|}{CT-Rate (16 tasks)} & \multicolumn{3}{c}{Rad-ChestCT (14 tasks)} \\ \cline{2-7} 
 & Mean & Mean & Mean & Mean & Mean & Mean \\
 & AUC & BAcc & wF1 & AUC & BAcc & wF1 \\ \hline
CT-CLIP~\cite{hamamci2024developing} & 73.3 & 66.9 & 70.8 & 63.3 & 59.9 & 64.7 \\
BIUD~\cite{cao2024bootstrapping} & 71.3 & 68.1 & 71.6 & 62.9 & 60.6 & 65.2 \\
Merlin~\cite{blankemeier2024merlin} & 72.8 & 67.2 & 70.9 & 64.4 & 61.9 & 66.3 \\
fVLM~\cite{shui2025large} & 77.8 & 71.8 & 75.1 & 68.0 & 64.7 & 68.8 \\
HLIP-RA~\cite{zhao2025scalable} & 77.7 & 71.4 & 74.7 & \underline{72.3} & \underline{68.4} & \underline{72.1} \\
HLIP-SA~\cite{zhao2025scalable}  & \underline{78.7} & \underline{72.4} & \underline{75.5} & 71.7 & 67.7 & 71.4 \\
\name~ & \textbf{83.2} & \textbf{76.5} & \textbf{78.9} & \textbf{74.7} & \textbf{69.8} & \textbf{73.2} \\ \bottomrule
\end{tabular}
}\cutabovecaption 
\caption{Results for zero-shot evaluation on single-sequence chest CT classification tasks, on the test split of CT-Rate~\cite{hamamci2024developing} as well as on an external validation dataset, Rad-ChestCT~\cite{draelos2021machine}. Mean zero-shot AUC, balanced accuracy, and weighted F1 scores are reported for each model, following \cite{zhao2025scalable} and \cite{shui2025large}. HLIP-RA and HLIP-SA refer to HLIP trained with raw annotations and summarized annotations respectively, as reported in~\cite{zhao2025scalable}. Per-task performance for HLIP and \name~ is in Table~\ref{tab:chestctfull} in Appendix~\ref{app:chestctfull}.}
\label{tab:chestct}
\cutbelowtable
\end{table}

\cutabovesubsection

\subsection{Remote Sensing}

\cutbelowsubsection
Itemized text supervision naturally occurs in remote sensing datasets such as RSICD~\cite{lu2017exploring}, which contains over 10K satellite remote sensing images of various geospatial locations with itemized text descriptions. We compare \name~against several CLIP-style baselines on a 30-way classification task. Since RSICD is a relatively small dataset, all compared models are initialized randomly without any pre-trained weights and only trained on RSICD training split for fairness of comparison. We include further details about implementation, baselines, and preprocessing in Appendix~\ref{app:rs-implementation}. We also found that applying Diverse Sampling (DS)~\cite{xiao2025flair} improves performance for RSICD (but not for medical tasks, see Appendix~\ref{app:ds-abla}), so DS is applied to the text items when training \name~and all baselines (except Vanilla CLIP and SigLIP). We show results in Table~\ref{tab:rsicd}, where \name~outperformed baselines in most metrics, demonstrating that \name~generalizes well to itemized text supervision outside of medical imaging.

\begin{table}[]
\centering
\scalebox{0.85}{
\begin{tabular}{l|ccc}
\toprule
 & \begin{tabular}[c]{@{}l@{}}Mean Rank\\ (out of 30)\end{tabular} & \begin{tabular}[c]{@{}l@{}}Top-1\\ Acc\end{tabular} & \begin{tabular}[c]{@{}l@{}}Top-5\\ Acc\end{tabular} \\ \hline
Vanilla CLIP~\cite{radford2021learning} & 5.08 & 34.2 & 74.0 \\
Vanilla SigLIP~\cite{zhai2023sigmoid} & 4.79 & 38.2 & 75.7 \\
DreamLIP~\cite{zheng2024dreamlip} & 4.99 & 32.2 & 67.6 \\ 
Multi-positive SigLIP~\cite{xiao2025flair} & 4.76 & \textbf{46.3} & 75.4 \\
FLAIR~\cite{xiao2025flair} & \underline{4.10} & 41.8 & \underline{76.1} \\
\name~(Ours) & \textbf{3.76} & \underline{46.2} & \textbf{78.7} \\ \bottomrule
\end{tabular}
}\cutabovecaption
\caption{ Result of zero-shot 30-way classification on remote sensing images (test split of RSICD~\cite{lu2017exploring} dataset). All models are randomly initialized and trained only with RSICD data.}
\label{tab:rsicd}
\cutbelowtable
\end{table}

\cutabovesubsection

\subsection{Proof of concept experiment: synthetic itemized text supervision for natural images}

\cutbelowsubsection

\begin{table}[]
\scalebox{0.8}[0.8]{
\begin{tabular}{@{\hspace{0pt}}l@{\hspace{1pt}}|@{\hspace{1pt}}r@{\hspace{2pt}}r@{\hspace{2pt}}r@{\hspace{2pt}}r|r@{\hspace{2pt}}r@{\hspace{2pt}}r@{\hspace{2pt}}r@{\hspace{0pt}}}
\toprule
 & \multicolumn{4}{c|}{MSCOCO} & \multicolumn{4}{c}{FLICKR} \\
 & I@1 & I@10 & T@1 & T@10 & I@1 & I@10 & T@1 & T@10 \\ \hline
Vanilla CLIP~\cite{radford2021learning} & 1.3 & 6.6 & 2.3 & 10.4 & 3.1 & 11.0 & 4.7 & 18.4 \\
Vanilla SigLIP~\cite{zhai2023sigmoid} & 2.1 & 9.9 & 3.5 & 15.6 & 4.7 & 16.5 & 6.9 & 25.7 \\
DreamLIP~\cite{zheng2024dreamlip} & 4.0 & 18.2 & 6.0 & 23.2 & 9.6 & 30.0 & 13.6 & 39.2 \\
Multi-positive SigLIP~\cite{xiao2025flair} & 4.0 & 18.0 & 6.0 & 23.3 & 9.2 & 30.3 & 11.9 & 36.9 \\
FLAIR~\cite{xiao2025flair} & \underline{5.4} & \underline{22.5} & \textbf{8.3} & \underline{29.4} & \underline{11.9} & \underline{36.1} & \underline{16.7} & \underline{46.0} \\
\name~(Ours) & \textbf{6.1} & \textbf{24.1} & \underline{8.1} & \textbf{31.1} & \textbf{14.4} & \textbf{38.8} & \textbf{19.2} & \textbf{51.8} \\ \bottomrule
\end{tabular}
}
\cutabovecaption
\caption{Results for zero-shot evaluation on MSCOCO~\cite{lin2015microsoftcococommonobjects} and Flickr~\cite{young2014image}. All models are randomly initialized and trained with Itemized-cc0.3M data only.}
\label{tab:cc03m}
\cutbelowtable
\cutbelowtable
\end{table}

In addition to the domains with naturally itemized text supervision, we also aimed to evaluate if \name~works well for learning natural image representations. We create a synthetic set of itemized captions based on the first 10\% of data in CC3M-ReCap~\cite{zheng2024dreamlip} called Itemized-cc0.3M, by using LLMs to generate itemized captions from the original captions (see Appendix~\ref{app:curation} for curation details). The scale of the synthetic data roughly matches the other tested datasets (brain MRI and CT, on the order of $10^5$ data pairs). For fairness of evaluation, we train all models with fully random initialization on Itemized-cc0.3M data only, with DS enabled, and we evaluate on MS-COCO~\cite{lin2015microsoftcococommonobjects} and Flickr~\cite{young2014image} retrieval. The results are shown in Table~\ref{tab:cc03m}, and \name~outperformed all baselines. Implementation and evaluation details are in Appendix~\ref{app:image-impementation}.

Note that the synthetic itemized text supervision is significantly more challenging than the original multi-positive supervision, as the total amount of words per image available for training is over 10X less in Itemized-cc0.3M compared to CC3M-Recap. The goal of this experiment is to serve as a proof-of-concept that \name~is a better option compared to existing methods when only itemized text supervision is available.

\begin{figure*}
\centering
\includegraphics[width=0.99\linewidth]{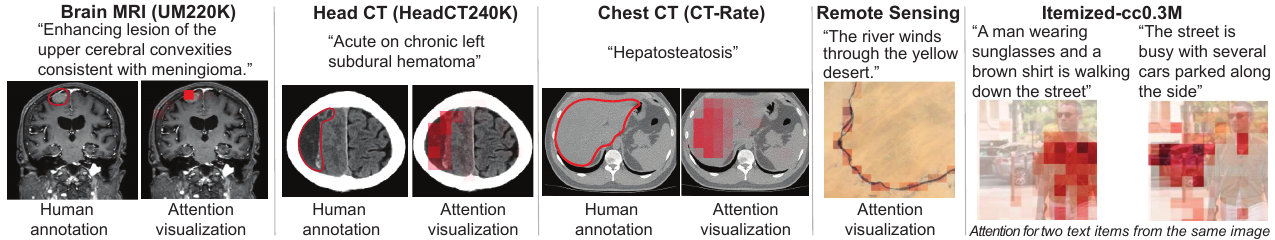}
\cutabovecaption
\caption{Examples of $\texttt{CrossAttn}$ visualization of text items from each of the 5 domains. On the 3D medical imaging domains, we only show attention on one slice from one sequence here due to space, but the attention visualization spans across all slices and sequences within each study, and we show expanded examples in Figure~\ref{fig:additionalvis2}. The human annotations were created by a medical professional to illustrate the ground truth region of interest with respect to the text item. }
\cutbelowtable
\cutabovecaption

\label{fig:visualizations}
\end{figure*}

\cutabovesection

\section{Analysis}

\cutbelowsection

\cutabovesubsection

\subsection{Item-level explainability}

\cutbelowsubsection

\begin{figure}
    \centering
    \includegraphics[width=1\linewidth]{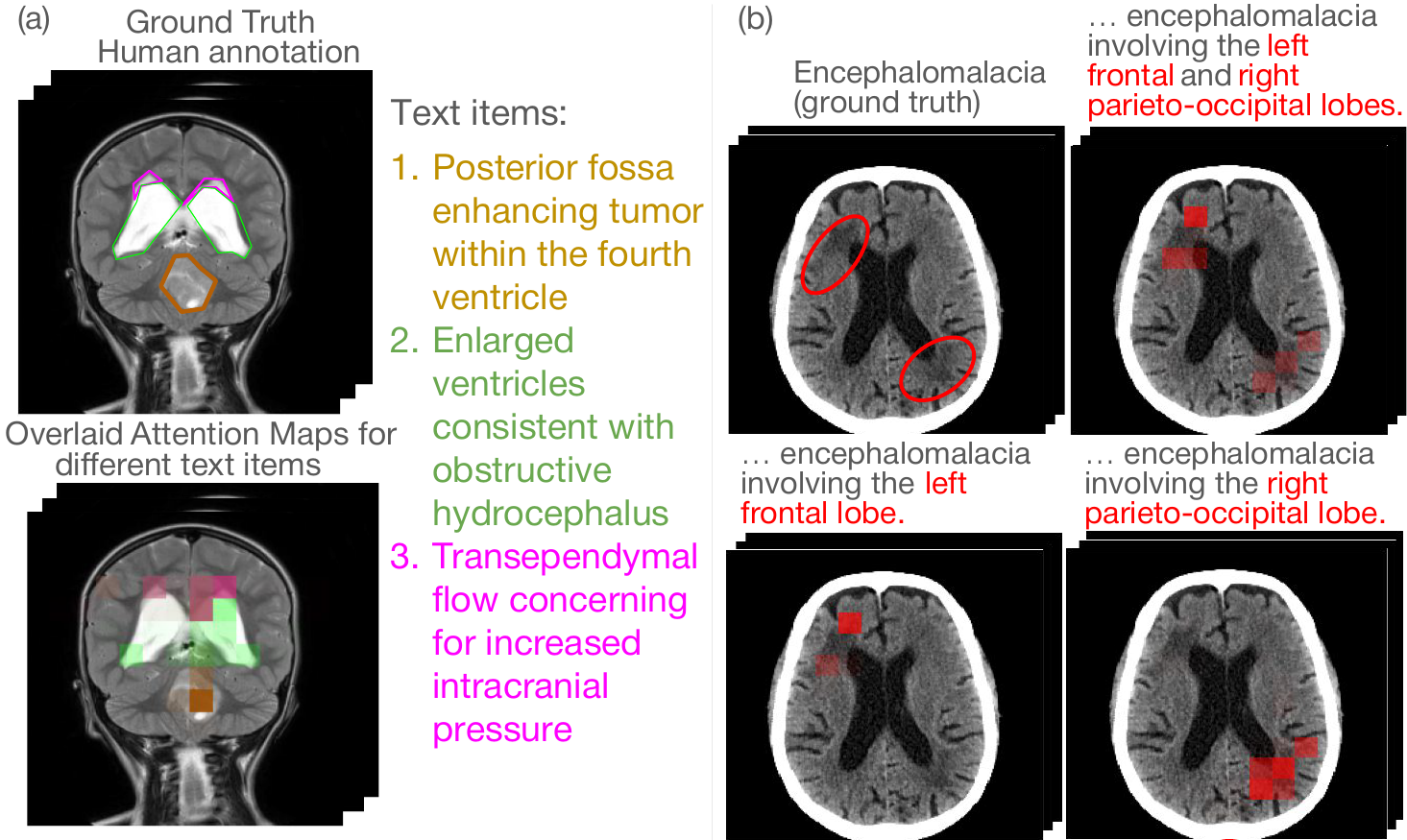}
    \cutabovecaption
    \cutabovecaption
    \cutabovecaption
    \caption{Example of \name's visualizations of different text items on the same brain MRI/head CT. \textbf{(a)} The overlaid attention maps for each of the 3 positive text items align with the ground truth regions. \textbf{(b)} A head CT shows encephalomalacia in two regions, and \name's visualizations can identify both and each one separately based on the text items, indicating \name's strong grounding and anatomical awareness without explicit anatomical supervision.}
    \label{fig:anat}
    \cutbelowtable
    \cutbelowtable
    \cutbelowtable
\end{figure}

\name-trained models are naturally explainable through attention scores from $\texttt{CrossAttn}$ module, similar to~\cite{xiao2025flair}. Text item cross-attention provides a direct method to evaluate if \name's representations are grounded. Given encoded visual tokens $vp$ and text item $t$, \name's explanations can be generated by taking the attention scores from $\texttt{CrossAttn}(t,vp,vp)$ and averaging across all attention heads. We include some visualization examples in Figure~\ref{fig:visualizations}, with additional examples in Appendix~\ref{app:visualization}. Explainability and grounding are especially essential in medical settings. One key advantage of \name~is that, while existing works like Prima~\cite{lyu2025learning} and HLIP~\cite{zhao2025scalable} can only explain classification logits through LIME~\cite{ribeiro2016lime} or attention map visualization, \name~can generate a visualization from natural language input, which enables verification of the model's understanding of fine-grained details. For example, in Figure~\ref{fig:anat}(a), \name~visualizes 3 different positive text items correctly on the same brain MRI. In Figure~\ref{fig:anat}(b), \name~generates visualizations that correctly point to two separate locations of the same pathology, confirming \name's accurate understanding of brain anatomy without explicit supervision from segmentation masks.

\cutabovesubsection

\subsection{Region-based text item retrieval}

\cutbelowsubsection

Under itemized text supervision, each text item often only refers to a specific small region within the image. We found that \name~has the capability to retrieve relevant text items given just a specific region of the image. Given encoded visual tokens $vp$ of the entire visual input, we take the subset of tokens $vp'$ within the region-of-interest, and retrieve text items $t$ from the text corpus that have the highest $TCSim(t,vp')$. We show an example in Figure~\ref{fig:differentiability}(a) and additional examples in Appendix~\ref{app:regionretrieval}.

\cutabovesubsection

\subsection{Better item differentiability via IIS}
\label{sec:differentiability}

\cutbelowsubsection

\begin{figure*}
    \centering
    \includegraphics[width=1.0\linewidth]{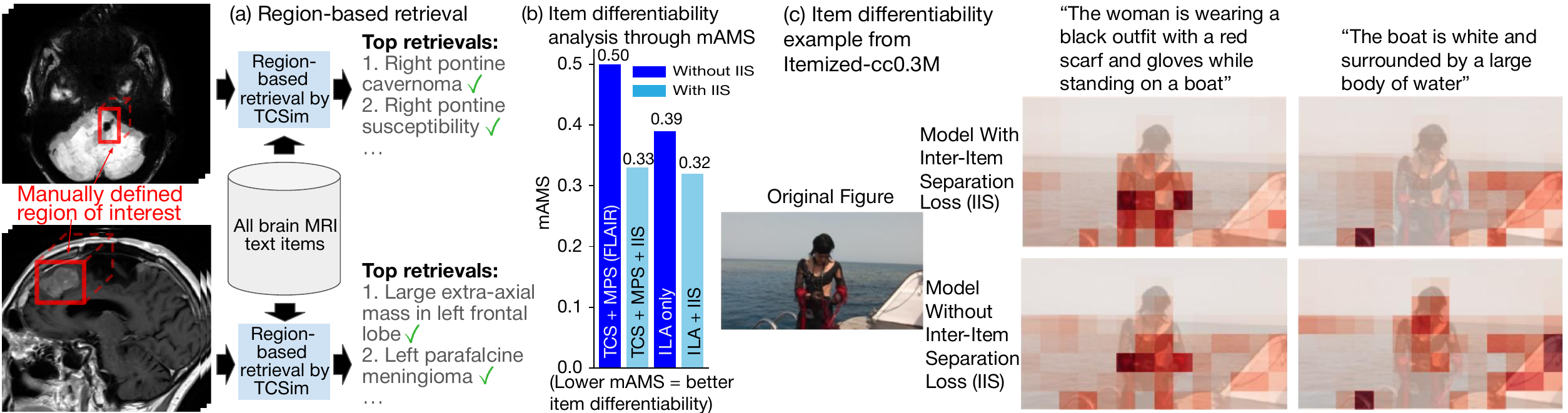}
    \cutabovecaption
    \cutabovecaption
    \cutabovecaption
    \caption{\textbf{(a)} Examples of region-based text item retrieval over brain MRIs. We manually provide a region mask. \name~can zero-shot retrieve the text items from the MRI text corpus that best describe the pathology within the region-of-interest (green check). In the bottom example, the retrieved text items describe the meningioma inside the region-of-interest, but not the resection cavity posterior to the region. \textbf{(b)} mAMS scores of several training settings with and without IIS on brain MRIs, evaluated on the prospective test set. We can see that when IIS is added to either FLAIR~\cite{xiao2025flair} (i.e. TCS+MPS) or ILA, mAMS goes down significantly, indicating that IIS forces the model to attend to different parts of the visuals for different text items. \textbf{(c)} An example from Itemized-cc0.3M that demonstrates item differentiability: our model trained with IIS is able to focus on the information mentioned in each text item, while the model trained without IIS has high attention on the woman even when the text item did not mention her.
    }
    \label{fig:differentiability}
    \cutbelowtable
\end{figure*}

\begin{table*}[]
\centering
\scalebox{0.83}{

\begin{tabular}{@{\hspace{0pt}}l@{\hspace{2pt}}|@{\hspace{1pt}}c@{\hspace{3pt}}c@{\hspace{3pt}}c@{\hspace{1pt}}c@{\hspace{2pt}}|@{\hspace{2pt}}c@{\hspace{3pt}}c@{\hspace{2pt}}|@{\hspace{2pt}}c@{\hspace{0pt}}}
\toprule
& \multicolumn{4}{@{\hspace{0pt}}c@{\hspace{2pt}}|@{\hspace{2pt}}}{\textbf{Ablations}} &
\multicolumn{2}{@{\hspace{0pt}}c@{\hspace{0pt}}|@{\hspace{2pt}}}{\textbf{Completeness}} &
\textbf{Segmentation} \\
 & Brain MRI & Chest CT & Remote Sensing & Itemized & Brain MRI & Head CT & Brain MRI \\
 & Pub-brain-5 & Rad-ChestCT & RSICD & -cc0.3M & Prospective & Prospective & BraTS2021 \\
  & \begin{tabular}[c]{@{}c@{}}12-task \\ mBAcc\end{tabular} & \begin{tabular}[c]{@{}c@{}}14-task\\ mAUC\end{tabular} & \begin{tabular}[c]{@{}c@{}}Top-1\\ Acc\end{tabular} & \begin{tabular}[c]{@{}c@{}}Flickr Retrieval\\ T@1\end{tabular} & \begin{tabular}[c]{@{}c@{}} Set\\ MLL*100\end{tabular} & \begin{tabular}[c]{@{}c@{}} Set\\ MLL*100\end{tabular} & \begin{tabular}[c]{@{}c@{}}Zero-shot Tumor\\ Segmentation mIoU\end{tabular} \\ \hline
Equal-weight TCS+MPS (FLAIR~\cite{xiao2025flair})& \multirow{1}{*}{78.7} & \multirow{1}{*}{73.9} & \multirow{1}{*}{41.8} & \multirow{1}{*}{16.7} & \multirow{1}{*}{31.12} & \multirow{1}{*}{41.00} & \multirow{1}{*}{11.4} \\ \hline
Text-conditioned SigLIP (TCS)  & 79.9  & 73.1 & 41.6 & 16.4 & 38.56 & 40.91 & 9.1 \\
+ Inter-Item Separation (IIS)  & 81.2 \textcolor{teal}{$^{+1.3}$} & 73.7 \textcolor{teal}{$^{+0.6}$} & 42.0 \textcolor{teal}{$^{+0.4}$} & 16.5 \textcolor{teal}{$^{+0.1}$} & 40.29 \textcolor{teal}{$^{+1.73}$} & 42.28 \textcolor{teal}{$^{+1.37}$} & 6.5 \textcolor{gray}{$^{-2.6}$} \\
+ Multi-Positive SigLIP (MPS) & 81.9 \textcolor{teal}{$^{+0.7}$} & \underline{74.2} \textcolor{teal}{$^{+0.5}$} & 42.8 \textcolor{teal}{$^{+0.8}$} & 16.5 $^{+0.0}$ & 39.43 \textcolor{gray}{$^{-0.86}$} & 41.71 \textcolor{gray}{$^{-1.57}$} & 16.1 \textcolor{teal}{$^{+9.6}$} \\
+ Upweight Worst Positive (UWP)  & 81.6 \textcolor{gray}{$^{-0.3}$} & 74.1 \textcolor{gray}{$^{-0.1}$} & 42.9 \textcolor{teal}{$^{+0.1}$} & 17.8 \textcolor{teal}{$^{+1.3}$} & \underline{42.96} \textcolor{teal}{$^{+2.50}$} & \underline{43.62} \textcolor{teal}{$^{+1.89}$} & \textbf{18.1} \textcolor{teal}{$^{+2.0}$} \\
+ Key Token Alignment (KTA)& \underline{82.7} \textcolor{teal}{$^{+1.1}$} & 74.1 $^{+0.0}$ & \underline{44.6} \textcolor{teal}{$^{+1.7}$} & \underline{17.9} \textcolor{teal}{$^{+0.1}$}  & 42.46 \textcolor{gray}{$^{-0.50}$} & 43.29 \textcolor{gray}{$^{-0.37}$} & 15.9 \textcolor{gray}{$^{-2.2}$} \\
+ TCS Masking (=\name) & \textbf{83.6} \textcolor{teal}{$^{+0.9}$} & \textbf{74.7} \textcolor{teal}{$^{+0.6}$} & \textbf{46.2} \textcolor{teal}{$^{+1.8}$} & \textbf{19.2} \textcolor{teal}{$^{+1.3}$} & \textbf{43.83} \textcolor{teal}{$^{+1.37}$} & \textbf{43.99} \textcolor{teal}{$^{+0.70}$} & \underline{17.5} \textcolor{teal}{$^{+2.6}$} \\ 
 \bottomrule
\end{tabular}

}

\cutabovecaption
\caption{Ablation study, representation completeness analysis, and zero-shot tumor segmentation results over incremental inclusion of objective components (ILA is divided into its 3 components: TCS, UWP, and TCS-Masking). In addition, we compare with FLAIR~\cite{xiao2025flair} objective (equal-weight TCS+MPS).}
\label{tab:ablations}
\cutbelowtable
\cutbelowtable

\end{table*}

Visual representations from itemized text supervision should be able to differentiate between two positive text items and attend to different visual regions when queried with different positive text items (see Figure~\ref{fig:anat}). IIS is the key objective that helps \name~achieve this. We design a metric to evaluate a model's item differentiability: mean Attention Map Similarity (mAMS), which measures average cosine similarity between visualization attention maps of two different positive items. We show mAMS of models trained with different objective combinations on brain MRI in Figure~\ref{fig:differentiability}(b), together with a qualitative example illustrating the difference in Figure~\ref{fig:differentiability}(c). We show an additional comparative example on brain MRI together with more general examples in Appendix~\ref{app:differentiability}. The results suggest that IIS strongly improves item differentiability.

\cutabovesubsection

\subsection{Ablation studies}

\cutbelowsubsection

We conduct ablation studies on each of our design choices on metrics from multiple domains. We present the results in ``Ablations" columns of Table~\ref{tab:ablations}, which shows that each component of \name~improves performance. In addition to the incremental addition of each component, we also show that \name~outperforms FLAIR~\cite{xiao2025flair} objective (equal-weight TCS+MPS).

\cutabovesubsection

\subsection{Representation Completeness}

\cutbelowsubsection

To measure how much \name~complies with the completeness criteria of itemized text supervision, we define a metric called \textbf{Mean Lowest Logit (MLL)}: for each visual representation $vp$ with corresponding text items $T={t_1,t_2,...,t_n}$, we calculate lowest logit as $\text{min}_{t\in T}TCSim(t,vp)$, and we average the lowest logit across all images in the testing datasets. A high MLL metric means most visual representations can be matched with all of their corresponding text items, thus satisfying the completeness criteria. We show MLL on brain MRI and head CT in the ``Completeness" columns of Table~\ref{tab:ablations}, where \name~has the highest MLL scores. Moreover, we observe that UWP, which was designed to improve representation completeness, indeed contributes the most in MLL scores; FLAIR, on the other hand, was designed for multi-positive supervision and thus has lower MLL scores.

\cutabovesection

\subsection{Zero-shot Tumor Segmentation}

\cutbelowsection

In addition to zero-shot classification, \name~can perform zero-shot segmentation via the attention heatmap between the encoded text vector of a description of the item to be segmented and encoded visual tokens, following~\cite{xiao2025flair}. We perform zero-shot segmentation over BraTS2021~\cite{baid2021brats}, which contains whole-study brain MRIs with ground truth tumor segmentation masks. We report mIoU of the zero-shot masks formed by top 30 visual tokens with the highest attention in the rightmost column of Table~\ref{tab:ablations}. The results indicate that strong segmentation performance is achieved only when both IIS and MPS are used. Implementation details and example visualizations are in Appendix~\ref{app:segmentation}.  

\section{Conclusion and Limitations}

We present \name, a language-supervised framework designed to learn complete and explainable visual representations from itemized text supervision, a naturally occurring but underexplored form of supervision where each image is described by multiple independent findings. \name~introduces innovations that enforce item differentiability and representation completeness, yielding interpretable and grounded attention maps and fine-grained alignment between text items and visual evidence. Across four real-world and one synthetic domain, \name~consistently delivers strong zero-shot performance and interpretable item-level grounding, advancing the broader goal of trustworthy vision–language learning.

Despite its generality, our exploration remains bounded by data scale and compute resources. We did not yet examined billion-scale pretraining or the effects of initializing from powerful pretrained vision–language models. Likewise, the approach depends on the quality and granularity of itemized text, which may vary across domains. Future work will explore large-scale fine-tuning, more efficient item-conditioned attention mechanisms, and longitudinal item reasoning across sequential or temporal datasets. We hope these findings inspire broader exploration of itemized text supervision as a natural and informative setting for training interpretable and semantically meaningful vision–language models.



\section*{Acknowledgements}

We would like to thank Karen Eddy (University of Michigan), Gary Laderach (University of Michigan), Brock Palen (University of Michigan) and Muhammad Bhalli (University of Michigan) for providing technical support. David Hanauer (University of Michigan) for support with the University of Michigan Electronic Medical Record Search Engine (EMERSE). Anthony Rosenzweig  (University of Michigan) for scientific guidance.

This work was supported by the following National Institute of Health (NIH) funding sources: K12NS080223 (T.C.H.). This work was supported by the Chan Zuckerberg Initiative (CZI), Frankel Institute for Heart and Brain Health (T.C.H.), the Mark Trauner Brain Research Fund, the Zenkel Family Foundation (T.C.H.), Ian’s Friends Foundation (T.C.H.) and the UM Precision Health Investigators Awards grant program (T.C.H.). The funders had no role in study design, data collection and analysis, decision to publish or preparation of the manuscript.

This research was also supported, in part, through computational resources and services provided by Advanced Research Computing, a division of Information and Technology Services at the University of Michigan.

%% file: X_suppl.tex
\clearpage
\setcounter{page}{1}
\maketitlesupplementary

\renewcommand{\thesection}{\Alph{section}}
\setcounter{section}{0}

\section{Expanded related works}
\label{app:rel}

\subsection{Visual representation learning with language supervision}

CLIP~\cite{radford2021learning} enabled visual representation learning from language supervision at large scale, using a contrastive objective to match image representation with text representation. Since then, many methods have been proposed to improve various aspects of CLIP training: SigLIP~\cite{zhai2023sigmoid} proposes an alternative contrastive loss formulation that enables better compute distribution and grants more flexibility in objective design, several approaches aim to combine the strength of self-supervised learning into CLIP objectives~\cite{mu2021slip,dong2023maskclip,jose2024dinov2meetstextunified}, and others aim to improve CLIP model's awareness of compositions by making the text in negative pairs more difficult~\cite{yuksekgonul2023when,rosch2024enhancing,liu2024empirical}. 

Recently, there has been an increasing demand for visual representations to capture more fine-grained details, and many works have tackled this through enforcing more fine-grained alignment between visual and text modalities:  PyramidCLIP~\cite{gao2022pyramidclip} attempts to perform alignment from global level to local level through a hierarchically designed objective, SoftCLIP~\cite{gao2024softclip} uses soft assignments to allow many-to-many mappings during training, and CoCa~\cite{yu2022coca} creates an additional caption-generation objective that uses a cross-attention layer to pool the local image tokens for caption generation, thus enforcing a more fine-grained image-to-text alignment.

Another sequence of existing research aim to improve visual-language contrastive learning by collecting more diverse positive captions for each image. LaCLIP~\cite{fan2023improving} and VeCLIP~\cite{lai2024veclip} uses LLMs to rewrite existing captions into alternative forms, and more recent advancements in Multimodal large language models (MLLMs) allow generating synthetic captions directly from the images using MLLMs~\cite{zheng2024dreamlip,yang2023alip,liu2023mllms,hammoud2024synthclip}. There has also been explorations in how to best incorporate all of these additional captions into CLIP training most effectively: LaCLIP~\cite{fan2023improving} simply randomly selects one caption each step during regular CLIP training; multi-positive versions of CLIP and SigLIP objectives~\cite{zheng2024dreamlip,xiao2025flair} has been proposed where multiple positive captions can be used within a single step of training; and Llip~\cite{lavoie2024modeling} first proposed to use a cross attention between text and visual representations to allow training with diverse positive captions without forcing representations of different positive captions to be close to each other.

Several recent works aim to combine multi-positive caption training with low-level visual-language alignment to achieve high quality fine-grained visual representations. DreamLIP~\cite{zheng2024dreamlip} proposed a text-conditioned visual representation based on cross attention between subcaption encodings and local visual tokens (instead of a few mixture tokens in Llip) to allow localization of subcaptions to specific local tokens. FLAIR~\cite{xiao2025flair} modifies the training objective from DreamLIP with different negative pairs (TCS loss) and combines it with Multi-positive SigLIP. These works have made significant progress in learning explainable, fine-grained and localized visual representations from multiple positive long captions, but are not designed for itemized text supervision. \name~adapts FLAIR's TCS+MPS objective (adding UWP and masking to TCS to obtain ILA) and added additional objectives (IIS and KTA) with the aim of learning visual representations from itemized text supervision that are not only high-quality, explainable and localizable, but also satisfy the requirements of itemized text supervision (item differentiability and completeness criteria).  

\subsection{Visual representation learning in medical imaging with itemized captions}

In many medical imaging domains (such as brain MRI, brain CT and chest CT that we explored in this paper), the associated language supervision (i.e. human-written reports) are often naturally well itemized. There exists many works that aim to learn visual representations through CLIP-like objectives using reports as supervision, but they often just concatenate the itemized reports into a single positive caption (e.g. Prima~\cite{lyu2025learning} for brain MRI, CT-CLIP~\cite{hamamci2024developing} for chest CT, and HLIP~\cite{zhao2025scalable} for all 3 domains) and lose the opportunity to learn fine-grained alignment between each item and local visual tokens. fVLM~\cite{shui2025large} decomposes chest CT reports into one single caption per organ, and uses pre-obtained segmentation masks to align each anatomical region in the chest CT volume with each organ caption in the reports. Although fVLM achieves an organ-level alignment, it requires availability of ground truth segmentation masks and does not directly achieve visual token-level alignment. \name~aims to fully take advantage of the itemized properties of the reports, and directly achieves alignment between each text item and corresponding local visual tokens, thereby allowing more explainable and localizable representations. \name~also adds additional completeness-based objectives (e.g. UWP) to ensure that the learned objective satisfies the completeness criteria, which is especially important within medical domains, as failing to pick up any of the abnormalities mentioned in a text item could lead to misdiagnosis.

\section{Itemized-cc0.3M curation details}
\label{app:curation}

\begin{figure*}
    \includegraphics[width=\linewidth]{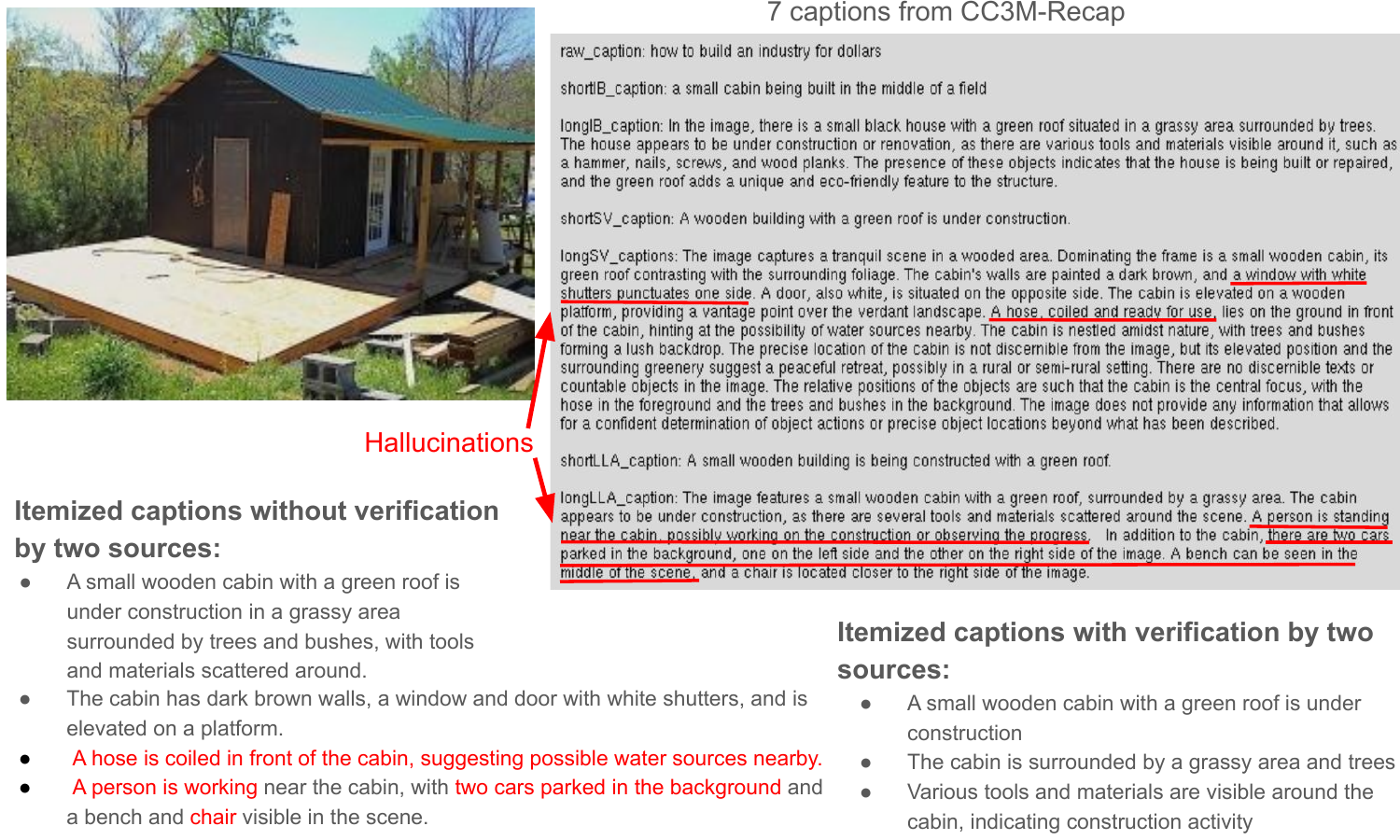}.
    \cutabovecaption
    \cutabovecaption
    \caption{An example image with 7 corresponding captions from CC3M-Recap~\cite{zheng2024dreamlip}. As shown in red, the VLM-generated captions contains quite a few pieces of hallucinated information. If we do not instruct GPT to perform two-source verification, some of the hallucinated information will go into the itemized dataset. With two-source verification, although some correct information is filtered out by the verification, it largely prevents any hallucinated information to become a completely wrong text item in the itemized captions.}
    \label{fig:itemizedexample1}
\end{figure*}

\begin{figure}
    \includegraphics[width=\linewidth]{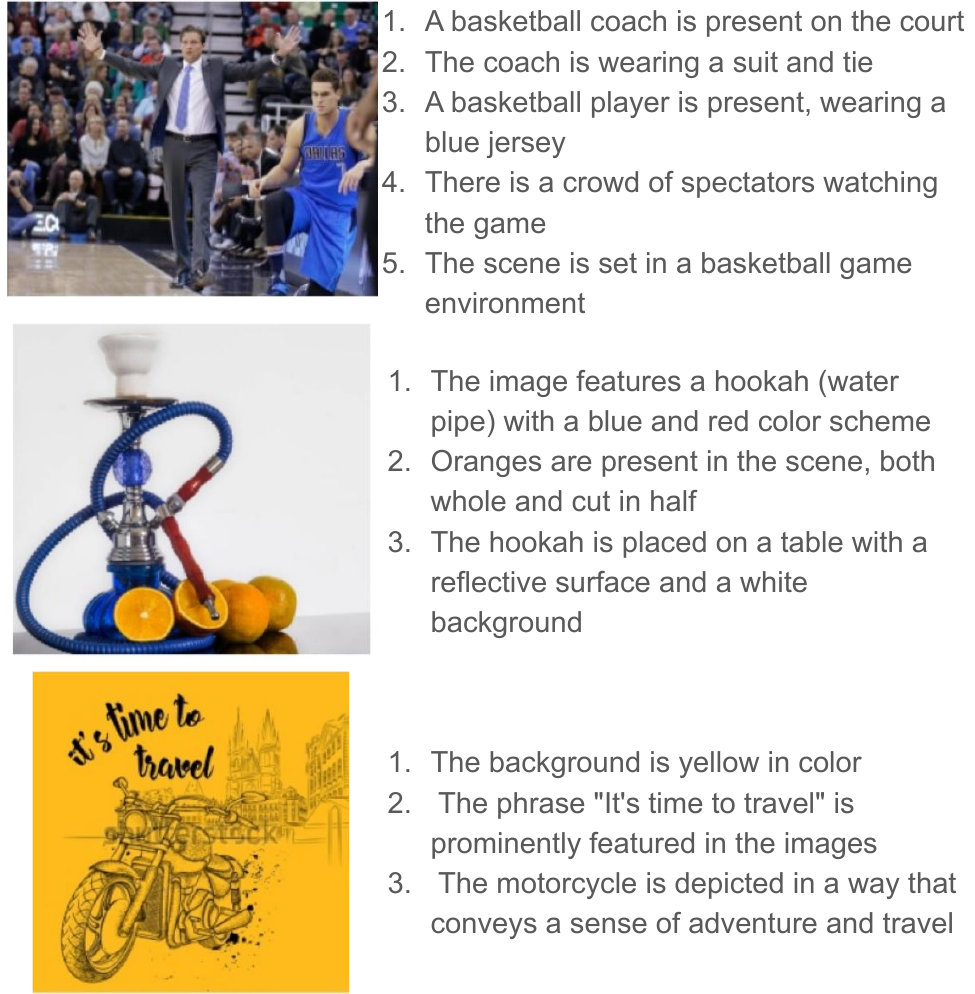}
    \cutabovecaption
    \cutabovecaption
    \caption{Additional caption examples from Itemized-cc0.3M.}
    \label{fig:itemizedexample2}
\end{figure}

CC3M-Recap~\cite{zheng2024dreamlip} is a paired image-caption dataset based on CC3M~\cite{sharma2018conceptual}, where in addition to the 1 original caption per image from CC3M, CC3M-Recap includes an additional 6 captions (3 long, 3 short) generated automatically with 3 different large VLMs. Training with CC3M-Recap is a typical multi-positive supervision due to high information overlap across the 7 captions. We aim to create synthetic itemized text supervision from these captions by using GPT-4o to write itemized captions given all 7 captions from CC3M-Recap. 

Since most of the captions are generated by VLMs, there are often hallucinatory statements within the automatically generated captions. While having a small hallucinatory part of a long caption in a multi-positive supervision can generally be tolerated, it would be very problematic if the hallucinatory part becomes a standalone text item in our synthetic itemized captions. Therefore, we added a verification step to each piece of information being included in our rewritten itemized captions: we call a piece of information ``verified" only if it has occurred in at least 2 of 7 original CC3M-Recap captions (it is unlikely for two different VLMs to make the exact same hallucination), and we only allow verified information in our itemized captions.  We want our synthesized captions to have minimum information overlap across items, and together covering all verified information (but not unverified ones that only occurred once among the 7 captions from CC3M-Recap), and we prompted GPT-4o to do so with the following system prompt:

\begin{promptbox}

\small
You are a helpful assistant. You will be provided with a set of 7 captions, 3 long ones and 4 short ones (including a raw one). Your goal is to generate an itemized list of summaries of the captions. 
Your list of items must satisfy: (1) The list should only include all information that has occurred in at least 2 of the provided captions; and (2) there should be no duplicate information across different entries in the summarized list. Each entry should be a plain sentence.

Your output should first generate a draft list with at least 5 entries, and then iteratively revised the draft list until the two conditions above are satisfied. Provide reason for each round of revision. The most important things to check during revision is the presence of duplicate information as well as whether the source of each information has occurred in at least 2 of the 7 provided captions. You should list the source captions for each of your list entries in each round of revision, and list the source count. If there is less than 2 sources listed  for this entry, you should remove the entry in the next round of revision.

In addition, list the Primary objects and/or properties that are the focus of each entry. If you find that a part of an entry in the list has duplicate information (about object, attribute, or concept) with another, you should remove the information from this entry during revision, while keeping the remaining unduplicated information in the entry. Each piece of information should only appear once across all entries. In addition, if there exists any two entries that are very similar (i.e. describes the same subject's similar attributes), you should combine the information from those two entries into one. Each list entry should primarily focus on a different set of objects or attributes. Your final entries should contain all remaining details after revision.

When you are done revising, you must output your final list in the following format: $||$finallist$||$ $<$final entry 1$>$ $|$ $<$ final entry 2$>$ $|$ ...  
\end{promptbox}

We show a concrete example with the original 7 captions in Figure~\ref{fig:itemizedexample1}, together with GPT-generated itemized captions with and without the two-source verification. As shown in red, the original captions contain quite a few pieces of information that is inaccurate, and two-source verification is able to remove them from the itemized captions.

We perform the above GPT-4o rewriting on the captions of the first 300K images in CC3M-Recap to form the Itemized-cc0.3M dataset. We show a few examples of Itemized-cc0.3M dataset in Figure~\ref{fig:itemizedexample2}. Note that we do not intend to make any claims about whether itemized captions in Itemized-cc0.3M is superior to those in CC3M-Recap or not. Itemized-cc0.3M is heavily filtered, which means it contains much less hallucinations, but it also contains significantly less overall text compared to the captions in CC3M-Recap. The experiments conducted on Itemized-cc0.3M is only intended to show that, in a hypothetical situation where a natural image dataset with only itemized captions is provided, \name~outperforms alternative options.

\section{Implementation Details and Hyperparameters}

\subsection{Whole-study Brain MRI}
\label{app:mri-implementation}

\begin{figure*}
    \includegraphics[width=\linewidth]{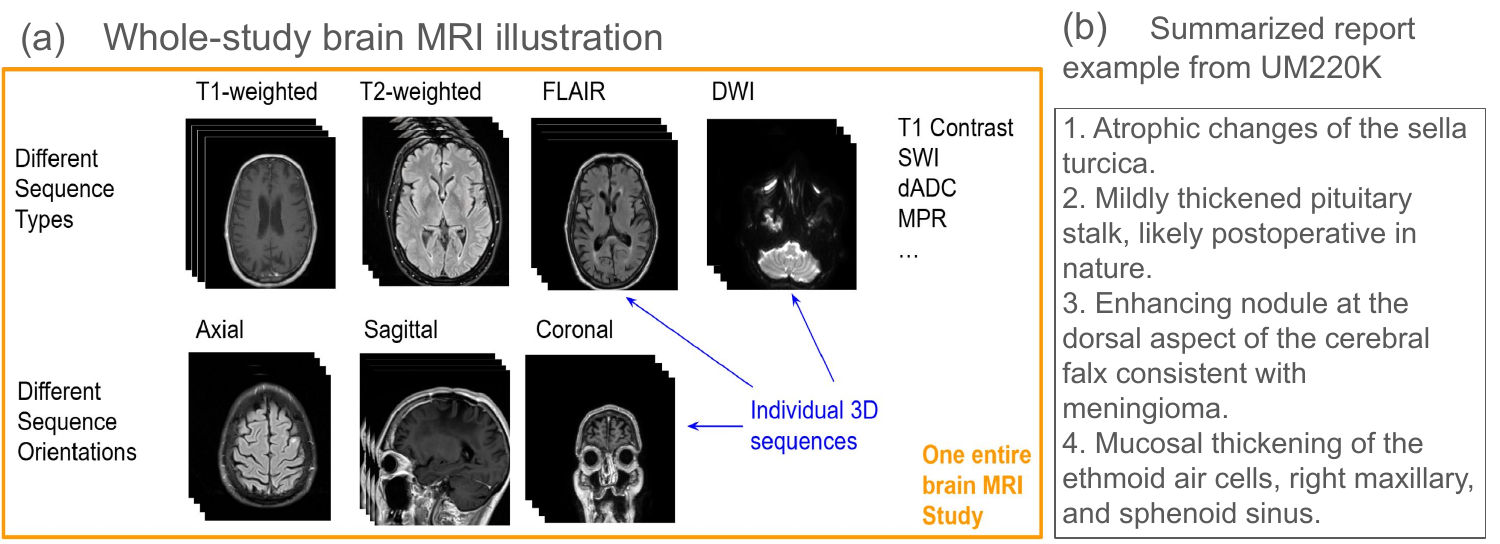}
    \caption{textbf{(a)}: Illustration of brain MRI whole-study. Each brain MRI whole-study contains multiple 3D sequences of different types and orientations. Each 3D sequence contains many 2D slices, which are stacked together to form a dense 3D volume of the brain. \textbf{(b)}: An example of summarized report from UM220K. Brain MRIs are a canonical example of itemized text supervision.}
    \label{fig:mriexample}
\end{figure*}

Brain MRI is indispensable in modern neurosurgery and neuroradiology. When a patient takes a Brain MRI study, the result is a \textbf{whole-study} Brain MRI: a collection of many 3D MRI Sequences. Each 3D sequence consists of many 2D slices, and the 2D slices are ``stacked" together along the axis perpendicular to the slice plane to form a dense 3D volume depicting the brain. Each 3D sequence has a weighting type (e.g. T1-weighted, T2-weighted, T1-contrast, FLAIR, SWI, DWI, etc) and an orientation (axial, sagittal, coronal), and each whole-study usually contains many sequences with different types and orientations. Therefore, modeling whole-study brain MRI has been particularly challenging due to the sheer amount of information present in a single study: in UM220K~\cite{lyu2025learning}, a whole-study contains over 1600 2D slices on average. After a patient takes a brain MRI, the whole-study is sent to a trained radiologist for interpretation, and the radiologist will write a report associated with the entire study. The radiology report is typically well itemized, with each sentence describing distinct findings or impressions. The first large-scale study on whole-study brain MRI is Prima~\cite{lyu2025learning}, which trains a CLIP model between whole-studies and their corresponding reports (summarized by GPT into an itemized list of abnormalities only, included as part of UM220K dataset). We show an illustration of a whole-study as well as an example of summarized report in Figure~\ref{fig:mriexample}.

The data we use to train our model is UM220K~\cite{lyu2025learning}, same as Prima~\cite{lyu2025learning} and HLIP~\cite{zhao2025scalable}. It is the largest whole-study brain MRI dataset to the best of our knowledge, and it contains over 220K whole-studies with corresponding summarized radiologist reports. In addition, UM220K also contains an additional prospective test set of 30K whole-studies, separated temporally from the 220K training set by study date. All studies also have binary annotations over 52 diagnoses. 

HLIP~\cite{zhao2025scalable} is a model architecture designed to handle the large amount of data in a single radiology study via efficient hierarchical attention. Each 3D sequence is first reshaped to 48x256x256, then center-cropped to 48x224x224, and then is cut into 8x16x16 tokens. Each token is fed through a 3D CNN to be flattened into a 1D vector, and the sequence of tokens (as 1D vectors) is fed into a ViT-base with CLS token to be encoded. Our implementation of \name~uses the identical model architectures as HLIP for both visual encoder (ViT-base~\cite{vit} with HLIP hierarchical attention, pre-trained from``vit\_base\_patch16\_224.mae" from OpenCLIP~\cite{ilharco_gabriel_2021_5143773}) and text encoder (BiomedBERT~\cite{chakraborty-etal-2020-biomedbert}, from ``microsoft/BiomedNLP-BiomedBERT-base-uncased-abstract-fulltext" on Huggingface), with one minor deviation: Prima~\cite{lyu2025learning} found that including the names of each 3D sequence (e.g. ``AX\_T2", etc, available as part of the UM220K) slightly improves model performance, so we replaced the learnable sequence-order embeddings in HLIP visual encoder with the sequence name encoder from Prima (fully trainable) that encodes each sequence's name into an embedding. We call this modified architecture ``HLIP-SN". The modification is very lightweight and has negligible impact on computation and memory, but yields a small improvement in overall performance. To justify this design choice, we conduct an additional ablation study of implementing \name~with unmodified HLIP, and we show the results in Table~\ref{tab:sn} over the prospective test set. We can see that \name~still significantly outperforms HLIP even when implemented with identical architecture as HLIP, and using HLIP-SN further improves performance by a small amount. 

Another minor adjustment to \name~that we make for brain MRIs is called ``normal check": around 10\% of studies in UM220K training set have no abnormalities mentioned in the radiologist report, so the GPT-summarized report simply states ``Study is unremarkable". We simply mark these studies as ``normal" and we do not use text from one normal study and visual from another normal study as negative pairs in ILA, MPS and KTA. Applying normal check improves overall model performance, as shown in Table~\ref{tab:nc}.

\begin{table}[]
    \centering
    \scalebox{0.92}{
    \begin{tabular}{l|c}
    \toprule
         & mAUC over 52 tasks \\ \hline
    HLIP~\cite{zhao2025scalable}     & 83.7 \\
    \name~with unmodified HLIP & \underline{90.2} \\
    \name~with HLIP-SN & \textbf{90.5} \\ \bottomrule
    \end{tabular}
    }
    \caption{Zero-shot performance on prospective test set of UM220K. We see that \name~ significantly improves model performance over HLIP, whether implemented with unmodified HLIP or HLIP-SN. Implementing \name~ with HLIP-SN improves performance slightly.}
    \label{tab:sn}
\end{table}

\begin{table}[]
    \centering
    \scalebox{0.92}{
    \begin{tabular}{l|c}
    \toprule
         & mAUC over 52 tasks \\ \hline
    \name~without normal check & 89.6 \\
    \name~with normal check & \textbf{90.5} \\ \bottomrule
    \end{tabular}
    }
    \caption{Zero-shot performance on prospective test set of UM220K, comparing \name~with and without normal check. Normal check improves performance.}
    \label{tab:nc}
\end{table}

We train \name~on UM220K training data for 20 epochs, then we evaluate its zero-shot performance on UM220K prospective test set on 52 tasks. In addition, we also evaluate zero-shot performance of \name~ on Pub-brain-5~\cite{zhao2025scalable}, an evaluation benchmark for brain MRI foundation models formed by combining 5 publicly available datasets. Our zero-shot prompts for the prospective test set is identical to those used in the papers of Prima~\cite{lyu2025learning} and HLIP~\cite{zhao2025scalable}, where the best of 7 prompts is reported; and our zero-shot prompts for Pub-brain-5 is also identical to the ones from HLIP~\cite{zhao2025scalable} codebase. The baseline results presented on these evaluation benchmarks all came from~\cite{zhao2025scalable}.

\subsection{Whole-study Head CT}
\label{app:ct-implementation}

\begin{figure}
    \includegraphics[width=\linewidth]{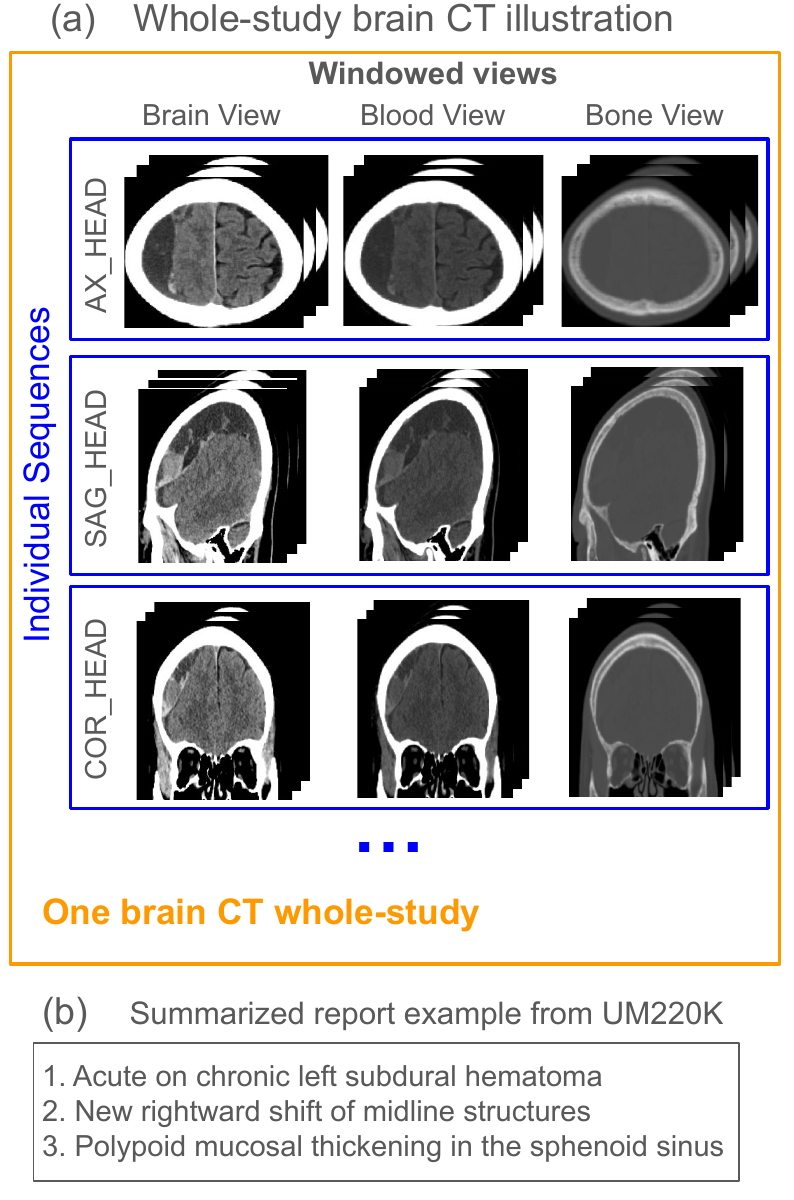}
    \cutabovecaption
    \cutabovecaption
    \caption{\textbf{(a)}: Illustration of brain CT whole-study. Each brain CT whole-study contains multiple 3D sequences of different orientations, and can be viewed in 3 different windows (brain, blood, bone). Each 3D sequence contains many 2D slices, which are stacked together to form a dense 3D volume of the brain. \textbf{(b)}: An example of summarized report from HeadCT240K.}
    \label{fig:ctexample}
\end{figure}

Brain CT is also a very important imaging technology for diagnosing brain diseases, especially hemorrhages and traumatic injuries. Similar to brain MRI, a brain CT whole-study also contains many 3D CT sequences of different types, where each sequences contains many 2D slices that are stacked together to form dense 3D structures. In addition, each 3D CT sequence is presented in 3 different views (brain, blood, bone), each obtained by setting a different window level and window width with respect to the original CT scanner HU and emphasizes different kinds of tissues. We illustrate this structure in Figure~\ref{fig:ctexample}(a). HeadCT240K~\cite{zhao2025scalable} is one of the largest datasets for whole-study brain CTs, with 240K CT whole-studies with paired reports. Like brain MRIs, after a patient takes a brain CT, the whole-study is sent to a radiologist for interpretation, and the radiologist will write a report associated with the entire study. The radiologist reports are already quite itemized, with each sentence usually talks about findings from unique regions. HeadCT240K also provides GPT-summarized radiologist reports that filtered out normal comments, so the summarized report is an itemized list of distinct anomalies found. We show an example of summarized report in HeadCT240K in Figure~\ref{fig:ctexample}(b). 

We again follow HLIP~\cite{zhao2025scalable} and preprocess the data by reshaping each 3D view into 48x256x256, center-cropping to 48x224x224, then cut into 8x16x16 tokens, and flatten with a 3D CNN layer before feeding into a ViT Base (the exact same preprocessing steps as brain MRIs above). The training setup is also identical to what we did for brain MRIs: we again use the HLIP-SN architecture for our model trained with~\name with the same pre-trained model initialization as HLIP and brain MRIs, and we apply normal check (see section~\ref{app:mri-implementation} for details). We treat each view of each sequence as a distinct sequence when inputting data into the HLIP encoder, following~\cite{zhao2025scalable}. The sequence names fed to HLIP-SN for each sequence contains both the name of the sequence and view window information for brain CTs (only sequence names are used for brain MRI, as brain MRI does not have different windowed views). We train \name~on the training set of HeadCT240K.

Like UM220K, HeadCT240K also has a prospective test set that is separated temporally by acquisition date from the training set. The prospective test set contains over 21K brain CT whole-studies with paired summarized reports, as well as diagnostic annotations for 83 diagnoses. We perform zero-shot evaluation on these diagnoses, using the same prompts as~\cite{zhao2025scalable}. In addition, we perform evaluation on 2 widely used benchmarks for evaluating brain CT foundation models: RSNA~\cite{flanders2020construction} has over 10K CT whole-studies (totalling over 25K 3D sequences) and annotations over 6 hemorrhagic diagnoses, and CQ500~\cite{chilamkurthy2018deep} has around 500 CT whole-studies and annotations on 10 diagnoses. We follow~\cite{zhao2025scalable} and~\cite{zhu20253d} to only report performance on 5 out of 6 diagnoses on RSNA. Same prompts were used to evaluate zero-shot classification on RSNA and CQ500 between \name and baselines (FullViT~\cite{zhao2025scalable} and HLIP~\cite{zhao2025scalable}).  We include our zero-shot prompts below:

\begin{promptbox}
\small

RSNA prompts for 5 diagnoses:

[``Intracranial hemorrhage", ``Intracranial hemorrhage with intraparenchymal hemorrhage", ``Intracranial hemorrhage with intraventricular hemorrhage", ``Intracranial hemorrhage with subarachnoid hemorrhage", ``Intracranial hemorrhage with subdural hemorrhage"]

CQ500 prompts for 10 diagnoses:

[``Intracranial hemorrhage", ``Intracranial hemorrhage with intraparenchymal hemorrhage", ``Intracranial hemorrhage with intraventricular hemorrhage", ``Intracranial hemorrhage with subdural hemorrhage", ``Intracranial hemorrhage with epidural hemorrhage", ``Intracranial hemorrhage with subarachnoid hemorrhage", ``Bleeding in left side", ``Bleeding in right side",``Midline shift", ``Mass Effect"]

\end{promptbox}

The baseline results reported on all brain CT benchmarks came from~\cite{zhu20253d} and \cite{zhao2025scalable}.

\subsection{Single-sequence Chest CT}
\label{app:chestct-implementation}

\begin{figure}
\centering
    \includegraphics[width=0.8\linewidth]{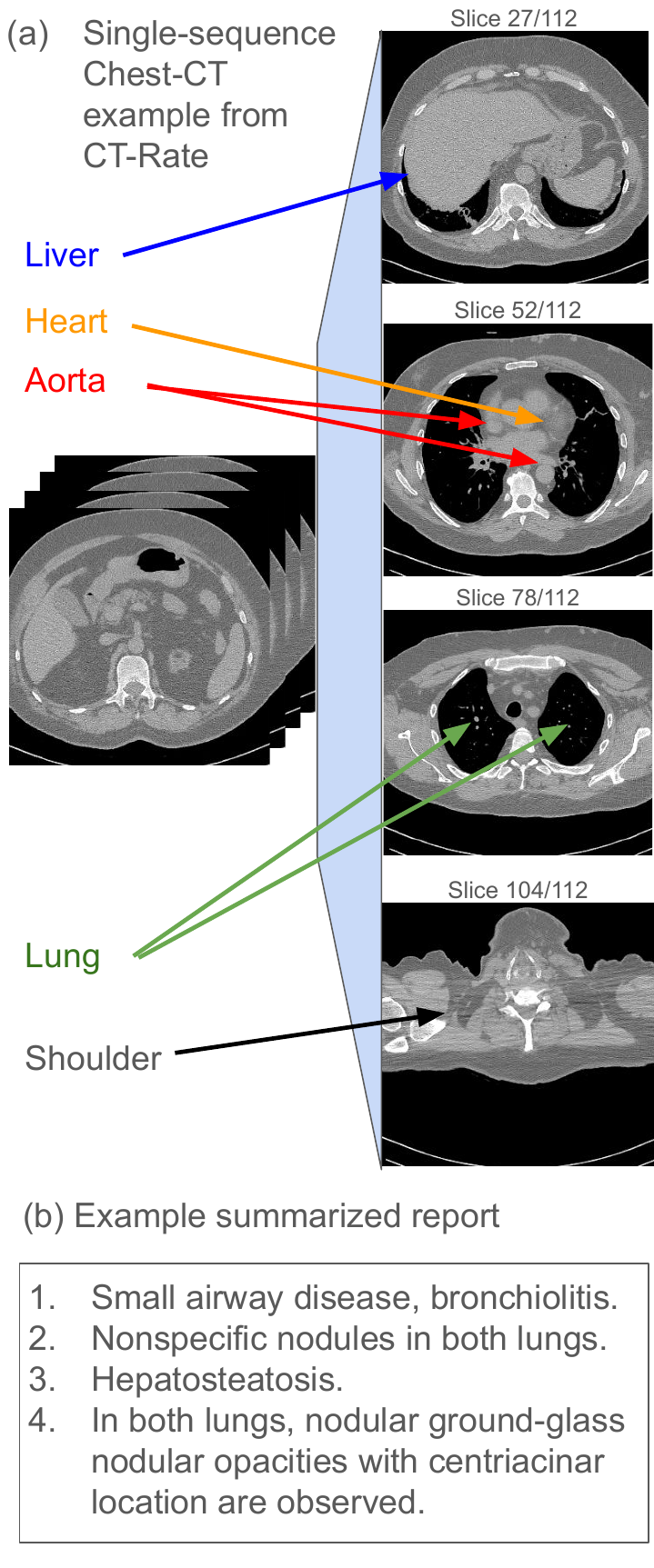}
    \cutabovecaption
    \cutabovecaption
    \caption{\textbf{(a)} An example of single-sequence chest CT from CT-Rate. As shown, it covers many different organs and regions such as shoulder, liver, heart, lung, and aorta. \textbf{(b)} An example summarized report for CT-Rate. The summarization simply removes any sentences describing no abnormal findings from the original report.}
    \label{fig:ctrateexample}
\end{figure}

The third medical imaging domain we include in our experiments is Single-sequence Chest CT, which contains a single 3D CT sequence taken over a patient's chest (from shoulder to upper abdomen). The dataset we use is CT-Rate~\cite{hamamci2024developing}, which contains 25.6K single-sequence chest CT with paired radiologist reports, and is split into training set and internal validation set (test set with 1.5K sequences). While the raw radiologist reports from CT-Rate can already be considered itemized (as each sentence usually refers to finding or no findings over a specific region of interest), existing works has typically further summarize the reports with LLMs: for example, fVLM~\cite{shui2025large} used LLM to summarize the reports into findings on 4 organs/regions (lung, heart, esophagus, aorta). We apply a rather simple approach: given a report (itemized by sentences), we simply asked GPT-5 to remove all items that do not describe an abnormality (e.g. items like ``no ... found" or ``no evidence of ..." or ``... appears normal"), similar to the summarized reports for brain MRI and CTs in Prima~\cite{lyu2025learning} and HLIP~\cite{zhao2025scalable}. We show an example chest CT sequence from CT-Rate as well as an example GPT-5 summarized report in Figire~\ref{fig:ctrateexample}.

When training \name, we again follow the same setup as HLIP~\cite{zhao2025scalable}: we first reshape each 3D sequence into 112x256x256 and center-crop to 112x224x224, then cut into 8x16x16 tokens which are flattened to 1D vectors through a 3D-CNN layer. The sequence of flattened tokens are then encoded by a ViT-Base model with CLS token. During training, we allow the center-crop to shift randomly by a few pixels as additional data augmentation. We initialize from the same pre-trained model as HLIP~\cite{zhao2025scalable}: the visual encoder initializes from ``vit\_base\_patch16\_224.mae" from OpenCLIP~\cite{ilharco_gabriel_2021_5143773}) and text encoder from BiomedVLP-CXR-BERT~\cite{boecking2022biomedvlp} (``microsoft/BiomedVLP-CXR-BERT-specialized" from Huggingface). We freeze the text encoder (following exact same practice as~\cite{zhao2025scalable}), and we employ normal check.

We follow existing works and evaluate our model on the internal validation split of CT-Rate~\cite{hamamci2024developing} (1.5K sequences, 16 diagnostic tasks) and an external validation dataset, Rad-ChestCT~\cite{draelos2021machine} test split (14 diagnostic tasks, 3.6K). Our zero-shot prompt is as follows (for both datasets, all tasks except ``calcification" are in CT-Rate internal validation and all tasks except ``mosaic attenuation pattern", ``Arterial wall calcification" and ``coronary artery wall calcification" are in Rad-ChestCT):

\begin{promptbox}
emphysema: ``findings consistent with emphysema"

atelectasis: ``findings consistent with atelectasis"

lung nodule: ``findings consistent with nodules or nodular density"

lung opacity: ``findings consistent with opacity"

pulmonary fibrotic sequela: ``findings consistent with pulmonary fibrotic sequela"

pleural effusion: ``findings consistent with pleural effusion"

peribronchial thickening: ``findings consistent with peribronchial thickening"

consolidation: ``findings consistent with consolidation"

bronchiectasis: ``findings consistent with bronchiectasis or bronchiectatic changes"

interlobular septal thickening: ``findings consistent with interlobular septal thickening"

cardiomegaly: ``findings consistent with cardiomegaly"

pericardial effusion: ``findings consistent with pericardial effusion"

coronary artery wall calcification: ``findings consistent with coronary artery wall calcification"

hiatal hernia: ``findings consistent with hiatal hernia"

arterial wall calcification: ``findings consistent with arterial wall calcification"

calcification: ``findings consistent with coronary artery wall calcification"
\end{promptbox}


\subsection{Remote Sensing}
\label{app:rs-implementation}

\begin{figure}
    \centering
    \includegraphics[width=0.8\linewidth]{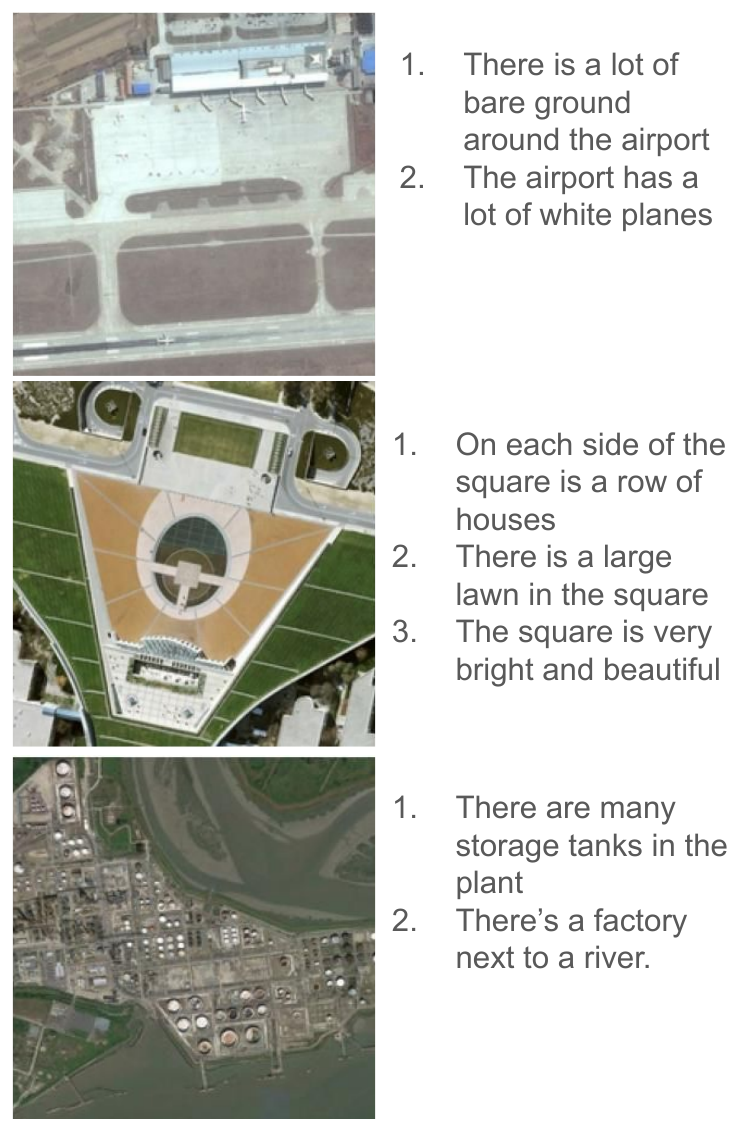}
    \caption{Examples of remote sensing images with paired deduplicated captions.}
    \label{fig:rsexample}
\end{figure}

RSICD~\cite{lu2017exploring} is a remote sensing dataset with 8.7K satellite image-caption pairs in the training split and 1.1K satellite image-caption pairs in the test split (we follow the data splitting from Huggingface dataset ``arampacha/rsicd"). Each satellite image in RSICD has 5 captions written by different annotators, each usually focuses on one specific object/property of the image. Sometimes there will be duplicates among the 5 captions for a single satellite image, so we use an LLM to deduplicate the captions (only removing duplicated captions, with no change to the wording to any remaining captions) for the training set and the resulting captions are well-itemized with low information overlap across items. We show example satellite images and their paired deduplicated caption items in Figure~\ref{fig:rsexample}.

We train~\name on the training split of RSICD. The model architecture and initialization is exactly the same as FLAIR~\cite{xiao2025flair}, with open-clip model config pasted below:

\begin{promptbox}

\{
    ``embed\_dim": 512,
    ``init\_logit\_bias": -10,
    ``vision\_cfg": \{
        ``image\_size": 224,
        ``layers": 12,
        ``width": 768,
        ``patch\_size": 16,
        ``output\_tokens": true
    \},
    ``text\_cfg": \{
        ``context\_length": 100,
        ``vocab\_size": 49408,
        ``width": 512,
        ``heads": 8,
        ``layers": 12
    \}
\}

\end{promptbox}

We train several CLIP-like baselines on the same training data: Vanilla CLIP~\cite{radford2021learning} and SigLIP~\cite{zhai2023sigmoid} are trained with all text items concatenated into one for each image, then apply the standard CLIP or SigLIP loss respectively; Multi-positive SigLIP~\cite{xiao2025flair} baseline is trained with MPS loss only; DreamLIP~\cite{zheng2024dreamlip} and FLAIR~\cite{xiao2025flair} are trained following each paper's original objective setups. We perform diverse sampling (DS)~\cite{xiao2025flair} as a data augmentation technique on all methods that takes in multiple positive items per image (DreamLIP, Multi-positive SigLIP, FLAIR, \name) as it significantly improves performance (see Appendix~\ref{app:ds-abla}).

We evaluate each method on the test split of RSICD. There is ground truth annotation of each image's category, out of 30 total categories (e.g. ``airport", ``dessert", ``dense residential area", ``industrial", ``port", etc). We perform zero-shot classification on which category each image belongs to out of the 30 categories. We report performance on top-1 accuracy, top-5 accuracy, and mean rank of the correct choice ranked by zero-shot logit of each category. We have 2 zero-shot prompts for each category, and we use the average of the 2 encoded text vectors when performing zero-shot. The 2 zero-shot prompts are as follows:

\begin{promptbox}
    ``There is $<$category name$>$"
    
    ``The $<$category name$>$"
\end{promptbox}

\subsection{Natural Images (Itemized-cc0.3M)}
\label{app:image-impementation}

We perform proof-of-concept experiment on Itemized-cc0.3M, a natural-image dataset consisting of 10\% of CC3M-Recap images with synthetically itemized text captions (curation process see Appendix~\ref{app:curation}, examples in Figure~\ref{fig:itemizedexample2}). The setups and training for both \name~and baselines are identical to those described in the previous subsection (for RSICD). We evaluate the models on 2 widely-used retrieval benchmarks: MSCOCO~\cite{lin2015microsoftcococommonobjects} and Flickr~\cite{young2014image}, and we follow the exact same zero-shot retrieval scripts as FLAIR~\cite{xiao2025flair}.

\subsection{Hyperparameters}

We report all hyperparameters used in training \name~on each of the 5 domains in Table~\ref{tab:hyperparameters}. 

\begin{table*}[]
\begin{tabular}{l|ccccc}
\toprule
 & Brain MRI & Brain CT & Chest CT & Remote Sensing & Itemized-cc0.3M \\ \hline
Weight of IIS loss ($\lambda_{\text{IIS}}$) & 1 & 1 & 1 & 1 & 0.1 \\
Weight of MPS loss ($\lambda_{\text{MPS}}$) & 0.01 & 0.1 & 0.1 & 2.0 & 0.1 \\
KTA loss key token rate ($K$) & 0.05 & 0.05 & 0.05 & 0.2 & 0.2 \\
Weight of KTA loss ($\lambda_{\text{KTA}}$) & 1 & 1 & 1 & 1.5 & 0.2 \\
TCS masking rate ($p_{\text{mask}}$) & 0.1 & 0.1 & 0.05 & 0.4 & 0.1 \\
Total Batch Size ($B$) & 256 & 256 & 512 & 1024 & 1024 \\
UWP upweighting factor ($w_{\text{uwp}}$) & 1.5 & 1.5 & 1.5 & 2 & 2 \\
Maximum number of text items per visual & 7 & 7 & 10 & 6 & 7 \\
Learning Rate & 0.000175 & 0.000175 & 0.0001 & 0.0003 & 0.0005 \\
Weight Decay & 0.2 & 0.5 & 0.5 & 1.5 & 0.8 \\
Optimizer & AdamW & AdamW & AdamW & AdamW & AdamW \\
Learning Rate Scheduler & Cosine & Cosine & Cosine & Cosine & Cosine \\
Scheduler beta1 & 0.9 & 0.9 & 0.9 & 0.9 & 0.9 \\
Scheduler beta2 & 0.98 & 0.98 & 0.98 & 0.98 & 0.98 \\
Total Epochs trained & 24 & 21 & 80 & 120 & 60 \\
Warmup Steps & 2000 & 2000 & 100 & 100 & 2000 \\
Initial SigLIP temperature ($\tau$) & 2.659 & 2.659 & 2.659 & 2.659 & 2.659 \\
Initial SigLIP bias ($b$) & -10 & -10 & -10 & -10 & -10 \\
Diverse Sampling Max Merged Num & - & - & - & 3 & 3 \\
Diverse Sampling Flag Probability & - & - & - & 0.5 & 0.5 \\ 
$\texttt{CrossAttn}$ number of heads & 8 & 8 & 8 & 8 & 8 \\ \bottomrule
\end{tabular}
\caption{Hyperparameters for training \name~on each domain.}
\label{tab:hyperparameters}
\end{table*}

\subsection{Zero-shot tumor segmentation}
 
\label{app:segmentation}

\begin{figure}
    \centering
    \includegraphics[width=\linewidth]{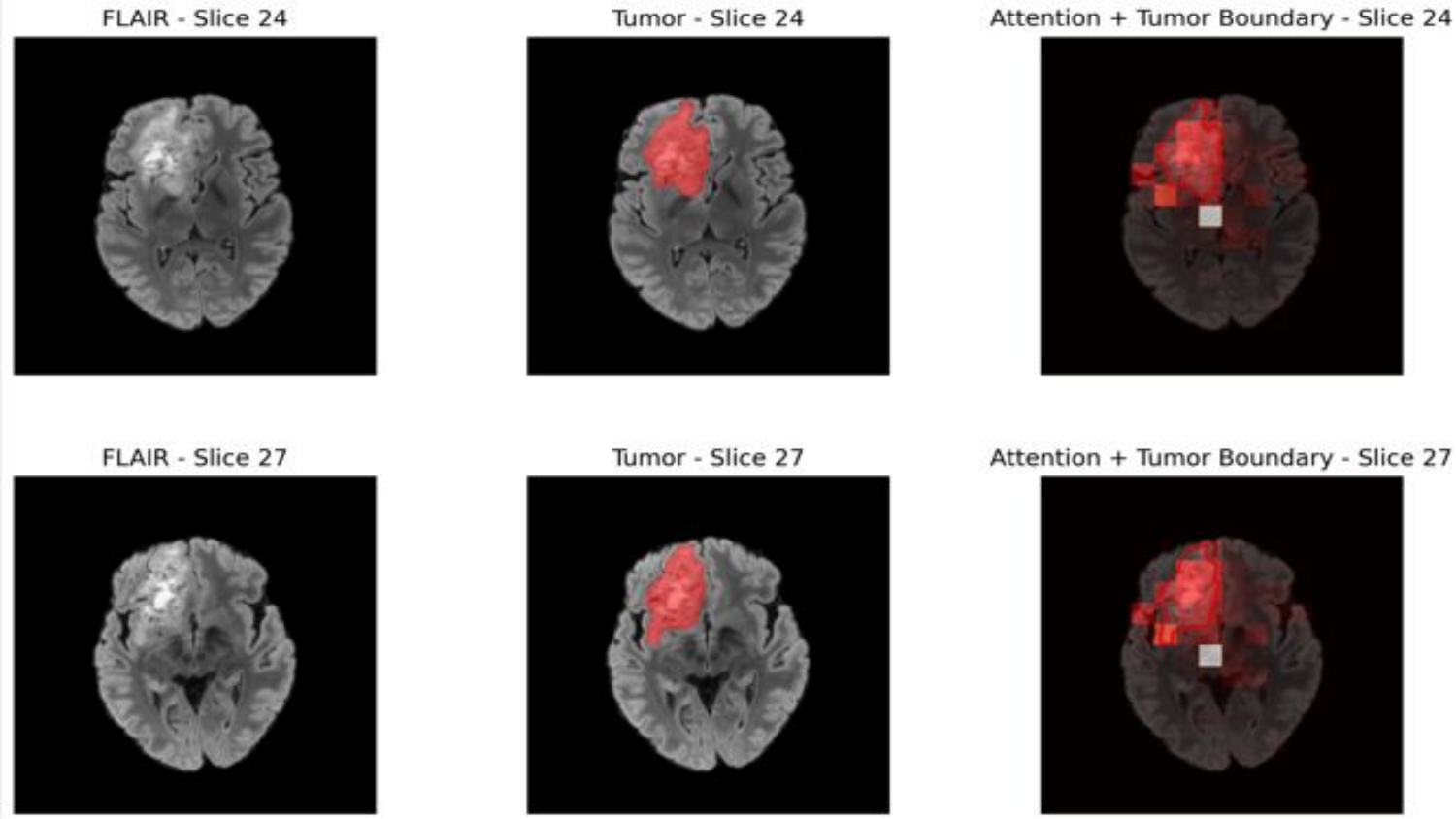}
    \cutabovecaption
    \cutabovecaption
    \caption{Example zero-shot segmentation result of \name~on BraTS2021~\cite{baid2021brats}. Two different slices from the same MRI sequence are shown. The middle column shows the ground truth segmentation mask, while the right column shows an overlay of the ground truth with our attention map. Our attention map resembles the ground truth segmentation mask.}
    \label{fig:segmentation}
\end{figure}

BraTS2021~\cite{baid2021brats} is a public dataset of curated brain MRI studies, where each study contains a fixed 4 axial sequences (T1, T2, FLAIR, T1CE). All sequences are skull-stripped for privacy preservation. All studies are from patients with Glioma, a type of brain tumor, and each FLAIR sequence has a ground truth segmentation mask. (Note that FLAIR here refers to a sequence type in brain MRI, not the visual-language pre-training method FLAIR~\cite{xiao2025flair}.) We take our model trained with \name~over UM220K and perform zero-shot segmentation by obtaining the attention map between each whole-study from BraTS2021 and the text prompt ``Glioma.", and take the top 30 visual tokens on the FLAIR sequence (out of 6x14x14=1176 total visual tokens for each sequence) as the zero-shot segmentation mask. Then, we compute IoU between zero-shot segmentation mask and the ground truth mask of the glioma, and average over 1251 studies in BraTS2021 to obtain the overall mIoU score reported in Table~\ref{tab:ablations}. We show an example from BraTS2021 in Figure~\ref{fig:segmentation}.

\section{Extended Results}

\subsection{Brain MRI UM220K prospective test set}

\begin{figure*}
    
    \centering
    \includegraphics[width=\linewidth]{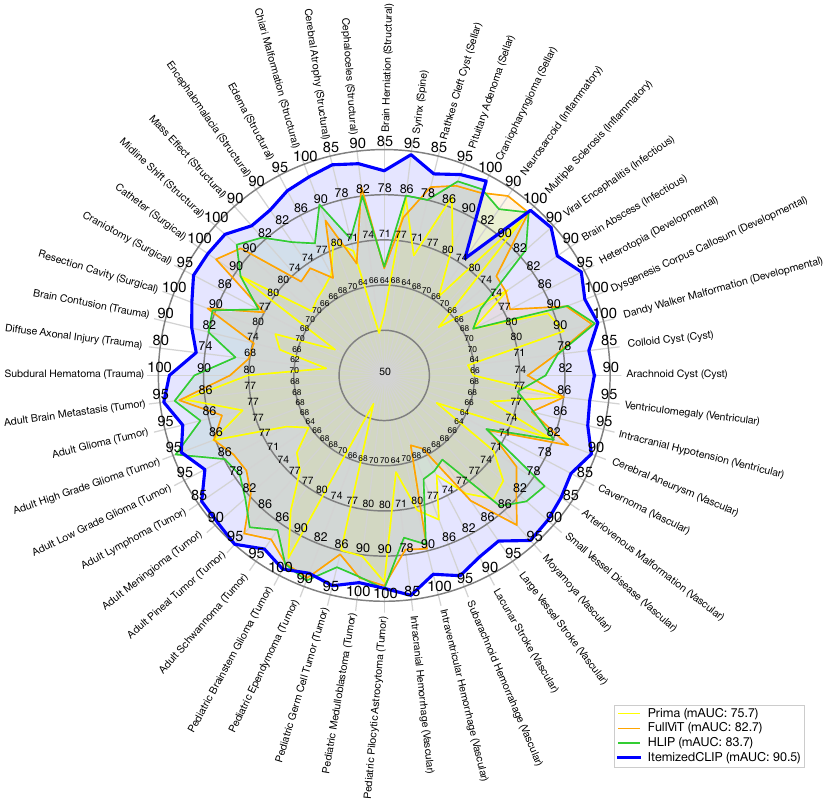}
    \caption{Radar plot on full results over all 52 tasks from Prima's prospective test set~\cite{lyu2025learning} (zero-shot AUC). The parenthesis after each diagnostic task is its category.}
    \label{fig:radar-mri}
\end{figure*}

We show the full task-wise results of 52 diagnostic tasks in UM220K prospective test set in Figure~\ref{fig:radar-mri}. The radar plot style is inspired by~\cite{lyu2025learning} and~\cite{zhao2025scalable}. \name~consistently outperforms baselines across an overwhelming majority of tasks.

\subsection{Brain CT prospective test set}

\begin{figure*}
    \centering
    \includegraphics[width=1.0\linewidth]{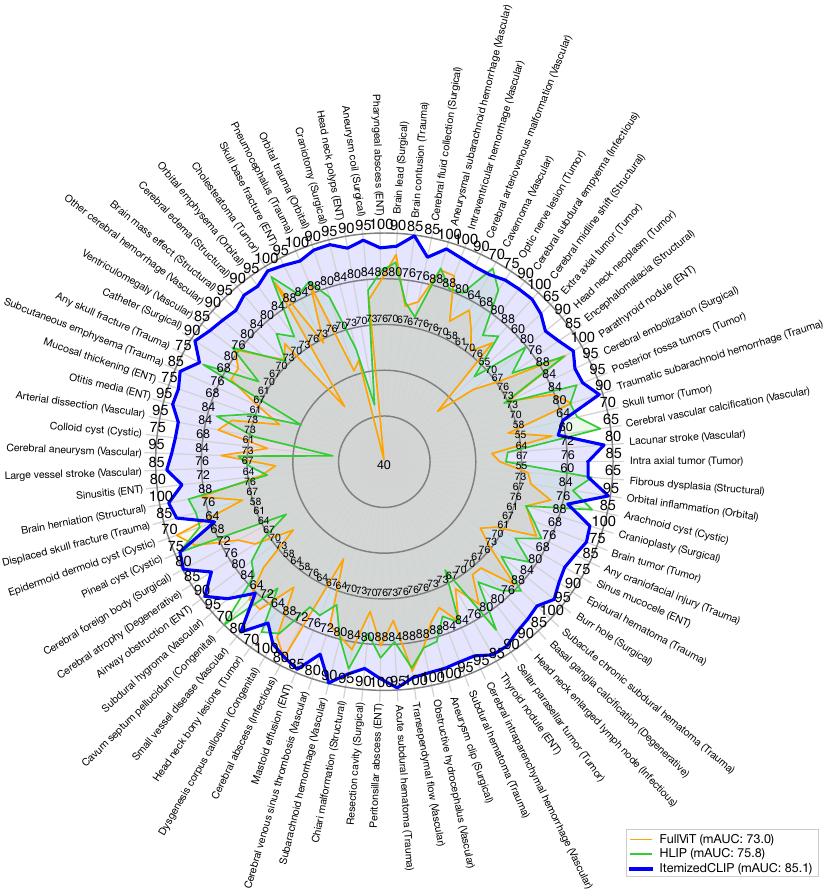}
    \caption{Radar plot on full results over all 83 tasks from HeadCT240K's prospective test set~\cite{zhao2025scalable} (zero-shot AUC). The radar plot style is inspired by~\cite{lyu2025learning} and~\cite{zhao2025scalable}. The parenthesis after each diagnostic task is its category.}
    \label{fig:radar-ct}
\end{figure*}

We show the full task-wise results of 83 diagnostic tasks in HeadCT240K's prospective test set in Figure~\ref{fig:radar-ct}. \name~consistently outperforms baselines across an overwhelming majority of tasks.

\subsection{Brain CT RSNA \& CQ500}
\label{app:rsnacq500}

We show expanded per-task results of brain CT evaluations on RSNA~\cite{flanders2020construction} in Table~\ref{tab:rsnafull} and on CQ500~\cite{chilamkurthy2018deep} in Table~\ref{tab:cq500full}. \name~consistently outperforms baselines in most tasks, and its zero-shot performance even outperforms linear probing performance of several recent CT foundation models (FM-HeadCT~\cite{zhu20253d},Google-CT~\cite{yang2024advancing} and Merlin~\cite{blankemeier2024merlin}).

\begin{table*}[]
\centering
\begin{tabular}{ll|ccccc|c}
\toprule
Models & Inference Type & \begin{tabular}[c]{@{}c@{}}ICH\\ any\end{tabular} & \begin{tabular}[c]{@{}c@{}}ICH\\ intraparenchymal\end{tabular} & \begin{tabular}[c]{@{}c@{}}ICH\\ intraventricular\end{tabular} & \begin{tabular}[c]{@{}c@{}}ICH\\ subarachnoid\end{tabular} & \begin{tabular}[c]{@{}c@{}}ICH\\ subdural\end{tabular} & \begin{tabular}[c]{@{}c@{}}Mean\\ AUC\end{tabular} \\ \hline
FullViT~\cite{zhao2025scalable} & Zero Shot & 79.0 & 82.6 & 92.7 & 82.8 & 79.2 & 83.3 \\
HLIP~\cite{zhao2025scalable} & Zero Shot & 81.5 & 88.2 & 91.4 & 84.1 & 83.4 & 85.7 \\
\name~(Ours) & Zero Shot & \underline{92.2} & \textbf{91.2} & \textbf{96.4} & \textbf{90.2} & \underline{87.4} & \textbf{91.5} \\
FM-HeadCT~\cite{zhu20253d} & Linear Probing & \textbf{92.6} & \underline{90.7} & \underline{95.6} & \underline{89.5} & \textbf{87.9} & 91.3 \\
Google-CT~\cite{yang2024advancing} & Linear Probing & 89.2 & 88.3 & 93.2 & 84.4 & 81.0 & 87.2 \\
Merlin~\cite{blankemeier2024merlin} & Linear Probing & 75.4 & 75.0 & 78.6 & 72.7 & 66.4 & 73.6 \\ \bottomrule
\end{tabular}
\caption{RSNA per-task AUC performance.}
\label{tab:rsnafull}
\end{table*}

\begin{table*}[]
\centering
\scalebox{0.85}{
\begin{tabular}{ll|c@{\hspace{5pt}}c@{\hspace{5pt}}c@{\hspace{5pt}}c@{\hspace{5pt}}c@{\hspace{5pt}}c@{\hspace{5pt}}c@{\hspace{5pt}}c@{\hspace{5pt}}c@{\hspace{5pt}}c|c}
\toprule
Models & Inference Type & ICH & IPH & IVH & SDH & EDH & SAH & BL-left & BL-right & MidlineShift & MassEffect & Mean AUC \\ \hline
FullViT~\cite{zhao2025scalable} & Zero Shot & 64.4 & 74.1 & 94.4 & 67.4 & 75.7 & 80.8 & 57.5 & 58.3 & 72.9 & 85.8 & 73.1 \\
HLIP~\cite{zhao2025scalable} & Zero Shot & 76.5 & \textbf{86.7} & \textbf{97.4} & \textbf{84.8} & \underline{88.0} & \underline{87.2} & \underline{72.0} & 75.6 & 75.9 & 86.8 & \underline{83.1} \\
\name~(Ours) & Zero Shot & \textbf{88.7} & \underline{86.4} & \textbf{97.4} & \underline{82.1} & \textbf{89.9} & \textbf{92.0} & \textbf{85.9} & \textbf{87.0} & \textbf{95.8} & \textbf{95.1} & \textbf{90.0} \\
FM-HeadCT~\cite{zhu20253d} & Linear Probing & 86.3 & 86.2 & \underline{95.9} & 65.3 & 65.2 & 80.7 & 71.4 & \underline{82.2} & 76.1 & \underline{90.7} & 80.0 \\
Google-CT~\cite{yang2024advancing} & Linear Probing & \underline{83.1} & 82.9 & 89.4 & 53.0 & 59.6 & 75.6 & 68.2 & 80.4 & \underline{90.1} & 78.4 & 76.1 \\
Merlin~\cite{blankemeier2024merlin} & Linear Probing & 58.9 & 61.6 & 49.8 & 55.0 & 68.9 & 50.3 & 48.3 & 54.9 & 51.4 & 51.4 & 55.1\\ \bottomrule
\end{tabular}
}
\caption{CQ500 per-task AUC performance.}
\label{tab:cq500full}
\end{table*}

\subsection{Chest CT CT-Rate \& Rad-ChestCT}

\label{app:chestctfull}

We show expanded per-task results of chest CT evaluations on CT-Rate~\cite{hamamci2024developing} and Rad-ChestCT~\cite{draelos2021machine} on \name~as well as two top-performing baselines, HLIP-RA and HLIP-SA~\cite{zhao2025scalable} in Table~\ref{tab:chestctfull}. \name~outperforms baselines in the majority of metrics, often by significant amounts.

\begin{table*}[]
\scalebox{0.9}{
\begin{tabular}{l|@{\hspace{2pt}}c@{\hspace{2pt}}c@{\hspace{2pt}}c@{\hspace{2pt}}c@{\hspace{2pt}}c@{\hspace{2pt}}c@{\hspace{2pt}}c@{\hspace{2pt}}c@{\hspace{2pt}}c@{\hspace{2pt}}|@{\hspace{2pt}}c@{\hspace{2pt}}c@{\hspace{2pt}}c@{\hspace{2pt}}c@{\hspace{2pt}}c@{\hspace{2pt}}c@{\hspace{2pt}}c@{\hspace{2pt}}c@{\hspace{2pt}}c}
\toprule
 & \multicolumn{9}{c|}{CT-Rate (16 tasks)} & \multicolumn{9}{c}{Rad-ChestCT (14 tasks)} \\
 & \multicolumn{3}{c|}{HLIP-RA} & \multicolumn{3}{c|}{HLIP-SA} & \multicolumn{3}{c|}{\name} & \multicolumn{3}{c|}{HLIP-RA} & \multicolumn{3}{c|}{HLIP-SA} & \multicolumn{3}{c}{\name} \\
 & AUC & BAcc & \multicolumn{1}{@{\hspace{2pt}}c@{\hspace{2pt}}|}{wF1} & AUC & BAcc & \multicolumn{1}{@{\hspace{2pt}}c@{\hspace{2pt}}|}{wF1} & AUC & BAcc & wF1 & AUC & BAcc & \multicolumn{1}{@{\hspace{2pt}}c@{\hspace{2pt}}|}{wF1} & AUC & BAcc & \multicolumn{1}{@{\hspace{2pt}}c@{\hspace{2pt}}|}{wF1} & AUC & BAcc & wF1 \\ \hline
Emphysema & 77.0 & 71.3 & \multicolumn{1}{@{\hspace{2pt}}c@{\hspace{2pt}}|}{73.8} & 77.3 & 69.2 & \multicolumn{1}{@{\hspace{2pt}}c@{\hspace{2pt}}|}{72.0} & \textbf{81.3} & \textbf{74.0} & \textbf{76.2} & 76.4 & 71.3 & \multicolumn{1}{@{\hspace{2pt}}c@{\hspace{2pt}}|}{72.5} & 74.0 & 67.7 & \multicolumn{1}{@{\hspace{2pt}}c@{\hspace{2pt}}|}{69.2} & \textbf{80.2} & \textbf{72.5} & \textbf{73.8} \\
Atelectasis & 70.2 & 64.1 & \multicolumn{1}{@{\hspace{2pt}}c@{\hspace{2pt}}|}{66.8} & 71.5 & 66.3 & \multicolumn{1}{@{\hspace{2pt}}c@{\hspace{2pt}}|}{68.5} & \textbf{76.8} & \textbf{68.3} & \textbf{70.6} & 63.1 & 62.0 & \multicolumn{1}{@{\hspace{2pt}}c@{\hspace{2pt}}|}{63.4} & 60.7 & 59.9 & \multicolumn{1}{@{\hspace{2pt}}c@{\hspace{2pt}}|}{61.4} & \textbf{67.6} & \textbf{62.8} & \textbf{64.4} \\
Lung nodule & 59.4 & 57.4 & \multicolumn{1}{@{\hspace{2pt}}c@{\hspace{2pt}}|}{57.5} & 61.3 & 59.0 & \multicolumn{1}{@{\hspace{2pt}}c@{\hspace{2pt}}|}{58.9} & \textbf{66.4} & \textbf{61.9} & \textbf{62.0} & 64.3 & \textbf{67.0} & \multicolumn{1}{@{\hspace{2pt}}c@{\hspace{2pt}}|}{\textbf{69.5}} & \textbf{65.9} & 63.9 & \multicolumn{1}{@{\hspace{2pt}}c@{\hspace{2pt}}|}{67.1} & \textbf{65.9} & 64.5 & 67.6 \\
Lung opacity & 79.9 & \textbf{76.6} & \multicolumn{1}{@{\hspace{2pt}}c@{\hspace{2pt}}|}{\textbf{76.6}} & 80.7 & 75.5 & \multicolumn{1}{@{\hspace{2pt}}c@{\hspace{2pt}}|}{75.5} & \textbf{81.1} & 76.1 & 76.3 & 66.2 & \textbf{61.8} & \multicolumn{1}{@{\hspace{2pt}}c@{\hspace{2pt}}|}{\textbf{61.8}} & 66.1 & 62.7 & \multicolumn{1}{@{\hspace{2pt}}c@{\hspace{2pt}}|}{62.5} & \textbf{66.4} & 61.7 & \textbf{61.8} \\
Pulmonary fibrotic sequela & 57.8 & 54.6 & \multicolumn{1}{@{\hspace{2pt}}c@{\hspace{2pt}}|}{57.0} & 59.1 & 55.9 & \multicolumn{1}{@{\hspace{2pt}}c@{\hspace{2pt}}|}{58.1} & \textbf{68.8} & \textbf{64.1} & \textbf{65.9} & 84.3 & 78.2 & \multicolumn{1}{@{\hspace{2pt}}c@{\hspace{2pt}}|}{81.1} & \textbf{84.8} & \textbf{81.9} & \multicolumn{1}{@{\hspace{2pt}}c@{\hspace{2pt}}|}{\textbf{84.0}} & 80.7 & 78.9 & 81.5 \\
Pleural effusion & 95.8 & 92.9 & \multicolumn{1}{@{\hspace{2pt}}c@{\hspace{2pt}}|}{93.3} & 95.5 & 93.0 & \multicolumn{1}{@{\hspace{2pt}}c@{\hspace{2pt}}|}{93.3} & \textbf{96.4} & \textbf{93.3} & \textbf{93.8} & \textbf{92.0} & \textbf{87.5} & \multicolumn{1}{@{\hspace{2pt}}c@{\hspace{2pt}}|}{\textbf{88.1}} & 89.3 & 84.1 & \multicolumn{1}{@{\hspace{2pt}}c@{\hspace{2pt}}|}{84.9} & 91.5 & 84.7 & 85.6 \\
Mosaic attenuation pattern & 72.2 & 62.5 & \multicolumn{1}{@{\hspace{2pt}}c@{\hspace{2pt}}|}{70.8} & 78.3 & 69.6 & \multicolumn{1}{@{\hspace{2pt}}c@{\hspace{2pt}}|}{76.3} & \textbf{86.4} & \textbf{80.9} & \textbf{84.6} & - & - & \multicolumn{1}{@{\hspace{2pt}}c@{\hspace{2pt}}|}{-} & - & - & \multicolumn{1}{@{\hspace{2pt}}c@{\hspace{2pt}}|}{-} & - & - & - \\
Peribronchial thickening & 73.2 & 67.4 & \multicolumn{1}{@{\hspace{2pt}}c@{\hspace{2pt}}|}{73.1} & 72.7 & 65.8 & \multicolumn{1}{@{\hspace{2pt}}c@{\hspace{2pt}}|}{71.7} & \textbf{80.5} & \textbf{74.1} & \textbf{78.4} & \textbf{64.6} & \textbf{65.9} & \multicolumn{1}{@{\hspace{2pt}}c@{\hspace{2pt}}|}{\textbf{73.4}} & 64.1 & 64.6 & \multicolumn{1}{@{\hspace{2pt}}c@{\hspace{2pt}}|}{72.5} & 64.5 & 60.4 & 69.1 \\
Consolidation & 87.7 & 79.1 & \multicolumn{1}{@{\hspace{2pt}}c@{\hspace{2pt}}|}{81.1} & 88.7 & 80.7 & \multicolumn{1}{@{\hspace{2pt}}c@{\hspace{2pt}}|}{82.2} & \textbf{90.8} & \textbf{82.9} & \textbf{84.1} & 79.9 & \textbf{74.8} & \multicolumn{1}{@{\hspace{2pt}}c@{\hspace{2pt}}|}{\textbf{78.2}} & 78.0 & 72.7 & \multicolumn{1}{@{\hspace{2pt}}c@{\hspace{2pt}}|}{76.6} & \textbf{80.4} & 74.0 & 77.7 \\
Bronchiectasis & 72.8 & 63.2 & \multicolumn{1}{@{\hspace{2pt}}c@{\hspace{2pt}}|}{70.2} & 72.9 & 70.5 & \multicolumn{1}{@{\hspace{2pt}}c@{\hspace{2pt}}|}{75.7} & \textbf{80.6} & \textbf{75.4} & \textbf{79.5} & \textbf{71.8} & \textbf{65.8} & \multicolumn{1}{@{\hspace{2pt}}c@{\hspace{2pt}}|}{\textbf{70.3}} & 70.1 & 63.8 & \multicolumn{1}{@{\hspace{2pt}}c@{\hspace{2pt}}|}{68.7} & 69.8 & 62.3 & 67.4 \\
Interlobular septal thickening & 82.3 & 73.9 & \multicolumn{1}{@{\hspace{2pt}}c@{\hspace{2pt}}|}{79.6} & 84.5 & \textbf{75.5} & \multicolumn{1}{@{\hspace{2pt}}c@{\hspace{2pt}}|}{\textbf{80.7}} & \textbf{85.2} & 73.7 & 79.5 & 74.6 & 65.6 & \multicolumn{1}{@{\hspace{2pt}}c@{\hspace{2pt}}|}{74.1} & 77.0 & 71.3 & \multicolumn{1}{@{\hspace{2pt}}c@{\hspace{2pt}}|}{78.4} & \textbf{82.4} & \textbf{78.3} & \textbf{83.3} \\
Cardiomegaly & 89.2 & 80.4 & \multicolumn{1}{@{\hspace{2pt}}c@{\hspace{2pt}}|}{83.6} & 91.9 & 81.9 & \multicolumn{1}{@{\hspace{2pt}}c@{\hspace{2pt}}|}{84.8} & \textbf{94.5} & \textbf{85.4} & \textbf{87.5} & 82.1 & 73.2 & \multicolumn{1}{@{\hspace{2pt}}c@{\hspace{2pt}}|}{78.0} & 83.5 & 74.9 & \multicolumn{1}{@{\hspace{2pt}}c@{\hspace{2pt}}|}{79.4} & \textbf{88.3} & \textbf{80.2} & \textbf{83.5} \\
Pericardial effusion & 83.2 & 72.1 & \multicolumn{1}{@{\hspace{2pt}}c@{\hspace{2pt}}|}{78.7} & 80.4 & 71.9 & \multicolumn{1}{@{\hspace{2pt}}c@{\hspace{2pt}}|}{78.5} & \textbf{85.9} & \textbf{80.2} & \textbf{84.4} & 64.6 & 60.2 & \multicolumn{1}{@{\hspace{2pt}}c@{\hspace{2pt}}|}{65.7} & 60.5 & 58.1 & \multicolumn{1}{@{\hspace{2pt}}c@{\hspace{2pt}}|}{63.8} & \textbf{66.6} & \textbf{63.9} & \textbf{68.8} \\
Coronary artery wall calcification & 87.0 & \textbf{78.8} & \multicolumn{1}{@{\hspace{2pt}}c@{\hspace{2pt}}|}{\textbf{79.8}} & 86.5 & 78.0 & \multicolumn{1}{@{\hspace{2pt}}c@{\hspace{2pt}}|}{79.1} & \textbf{87.1} & 78.4 & 79.6 & - & - & \multicolumn{1}{@{\hspace{2pt}}c@{\hspace{2pt}}|}{-} & - & - & \multicolumn{1}{@{\hspace{2pt}}c@{\hspace{2pt}}|}{-} & - & - & - \\
Hiatal hernia & 69.7 & 70.1 & \multicolumn{1}{@{\hspace{2pt}}c@{\hspace{2pt}}|}{74.3} & 68.8 & 64.3 & \multicolumn{1}{@{\hspace{2pt}}c@{\hspace{2pt}}|}{69.7} & \textbf{78.3} & \textbf{71.3} & \textbf{75.5} & 60.5 & 59.8 & \multicolumn{1}{@{\hspace{2pt}}c@{\hspace{2pt}}|}{67.0} & 60.0 & 56.4 & \multicolumn{1}{@{\hspace{2pt}}c@{\hspace{2pt}}|}{64.1} & \textbf{69.3} & \textbf{65.4} & \textbf{71.6} \\
Arterial wall calcification & 85.7 & 77.9 & \multicolumn{1}{@{\hspace{2pt}}c@{\hspace{2pt}}|}{78.7} & 89.0 & 81.4 & \multicolumn{1}{@{\hspace{2pt}}c@{\hspace{2pt}}|}{81.9} & \textbf{90.9} & \textbf{84.7} & \textbf{85.1} & - & - & \multicolumn{1}{@{\hspace{2pt}}c@{\hspace{2pt}}|}{-} & - & - & \multicolumn{1}{@{\hspace{2pt}}c@{\hspace{2pt}}|}{-} & - & - & - \\
Calcification & - & - & \multicolumn{1}{@{\hspace{2pt}}c@{\hspace{2pt}}|}{-} & - & - & \multicolumn{1}{@{\hspace{2pt}}c@{\hspace{2pt}}|}{-} & - & - & - & 68.1 & 64.7 & \multicolumn{1}{@{\hspace{2pt}}c@{\hspace{2pt}}|}{66.0} & 69.7 & 65.3 & \multicolumn{1}{@{\hspace{2pt}}c@{\hspace{2pt}}|}{66.6} & \textbf{72.9} & \textbf{68.0} & \textbf{69.3} \\ \hline
Mean & 77.7 & 71.4 & \multicolumn{1}{@{\hspace{2pt}}c@{\hspace{2pt}}|}{74.7} & 78.7 & 72.4 & \multicolumn{1}{@{\hspace{2pt}}c@{\hspace{2pt}}|}{75.5} & \textbf{83.2} & \textbf{76.5} & \textbf{78.9} & 72.3 & 68.4 & \multicolumn{1}{@{\hspace{2pt}}c@{\hspace{2pt}}|}{72.1} & 71.7 & 67.7 & \multicolumn{1}{@{\hspace{2pt}}c@{\hspace{2pt}}|}{71.4} & \textbf{74.7} & \textbf{69.8} & \textbf{73.2} \\ \bottomrule
\end{tabular}
}
\caption{Per-task zero-shot performance on Chest CT datasets (CT-Rate~\cite{hamamci2024developing} test set and Rad-ChestCT~\cite{draelos2021machine}), comparing \name~to two top-performing baselines (HLIP-RA and HLIP-SA~\cite{zhao2025scalable}).}
\label{tab:chestctfull}
\end{table*}

\subsection{Diverse Sampling analysis on different domains}
\label{app:ds-abla}

\begin{table*}[]
\centering
\begin{tabular}{l|ccc}
\toprule
 RSICD & \begin{tabular}[c]{@{}l@{}}Mean Rank\\ (out of 30)\end{tabular} & \begin{tabular}[c]{@{}l@{}}Top-1\\ Acc\end{tabular} & \begin{tabular}[c]{@{}l@{}}Top-5\\ Acc\end{tabular} \\ \hline
\name~without Diverse Sampling & 4.72 & 39.2  & 73.3 \\
\name~with Diverse Sampling & \textbf{3.86} & \textbf{46.2} & \textbf{78.7} \\ \bottomrule
\end{tabular}

\begin{tabular}{l|cccc|cccc}
\toprule
Itemized-cc0.3M & \multicolumn{4}{c|}{MSCOCO} & \multicolumn{4}{c}{Flickr} \\
 & I@1 & I@10 & T@1 & T@10 & I@1 & I@10 & T@1 & T@10 \\ \hline
\name~without Diverse Sampling & 4.5 & 20.0 & 6.0 & 25.5 & 10.2 & 33.3 & 14.2 & 43.0 \\
\name~with Diverse Sampling & \textbf{6.1} & \textbf{24.1} & \underline{8.1} & \textbf{31.1} & \textbf{14.4} & \textbf{38.8} & \textbf{19.2} & \textbf{51.8} \\ \bottomrule
\end{tabular}

\begin{tabular}{l|ccc|ccc}
\toprule
 \multirow{2}{*}{Chest CT}  & \multicolumn{3}{c|}{CT-RATE (16 tasks)} & \multicolumn{3}{c}{Rad-ChestCT (15 tasks)} \\
 & AUC & \multicolumn{1}{c}{Balanced ACC} & \multicolumn{1}{c|}{Weighted F1} & AUC & \multicolumn{1}{c}{Balanced ACC} & \multicolumn{1}{l}{Weighted F1} \\ \hline
\name~without Diverse Sampling & \multicolumn{1}{r}{\textbf{83.2}} & \textbf{76.5} & \textbf{78.9} & \multicolumn{1}{r}{\textbf{74.7}} & \textbf{69.8} & \textbf{73.0} \\
\name~with Diverse Sampling & \multicolumn{1}{r}{81.6} & 74.9 & 77.5 & \multicolumn{1}{r}{72.9} & 68.3 & 71.8 \\
\bottomrule
\end{tabular}

\caption{Comparison between applying diverse sampling versus no diverse sampling. Diverse sampling improves performance over remote sensing and natural image tasks, but does not help in medical imaging tasks such as Chest CT.}
\label{tab:ds}
\end{table*}

Diverse sampling (DS) is a data augmentation strategy on the text side developed in~\cite{xiao2025flair}. Diverse sampling randomly combines one or more text items together into one single longer text item. We found that applying diverse sampling significantly improves performance for 2D image domains (RSICD and Itemized-cc0.3M) when training with \name, but it does not work well for medical imaging domains and actually decreases performance of \name. We show the comparisons in Table~\ref{tab:ds}.

\section{Additional qualitative examples}

\subsection{Text item visualizations}
\label{app:visualization}

\begin{figure*}
    \centering
    \includegraphics[width=\linewidth]{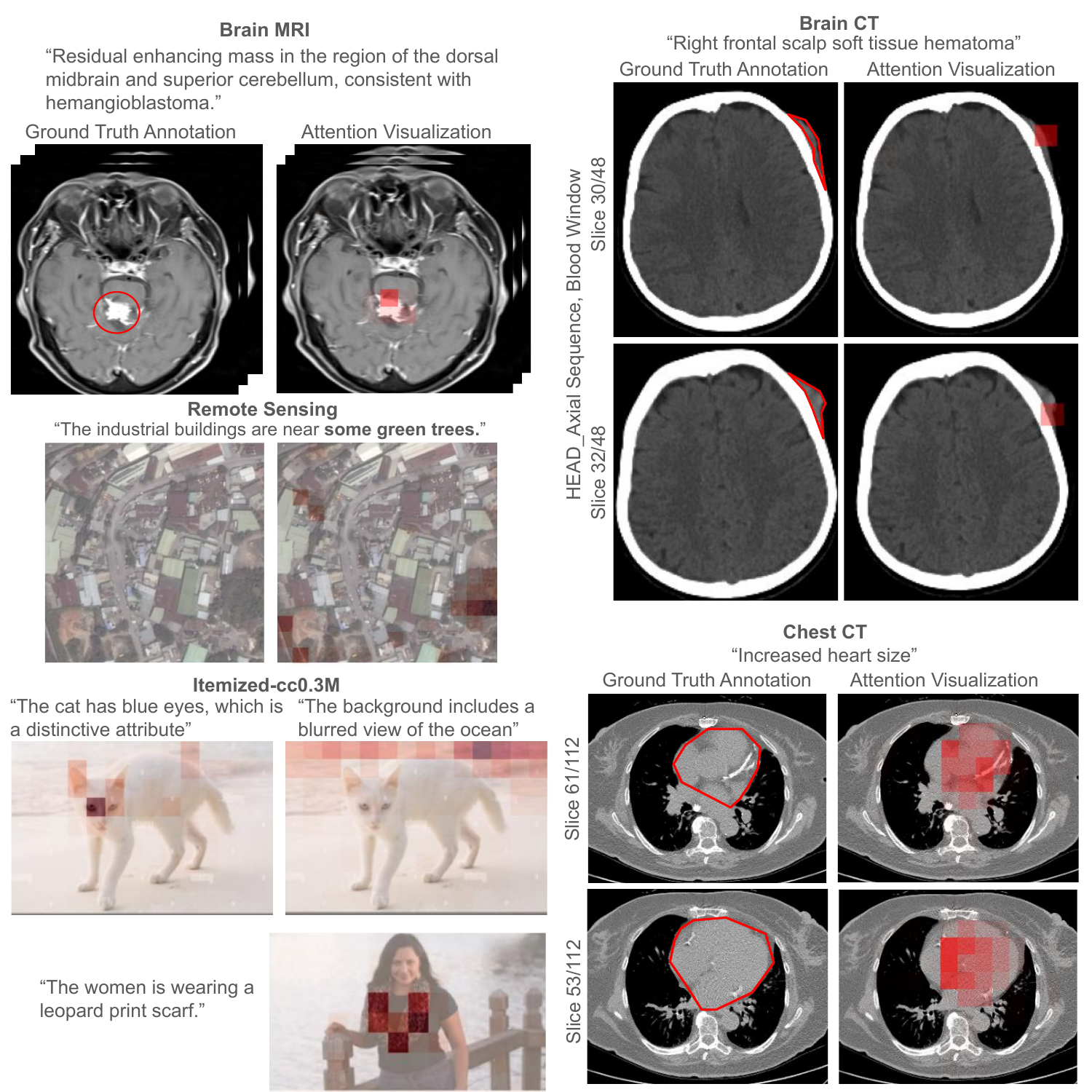}
    \caption{Additional text item attention visualization examples on all 5 evaluated domains.}
    \label{fig:additionalvis1}
\end{figure*}

\begin{figure*}
    \centering
    \includegraphics[width=0.9\linewidth]{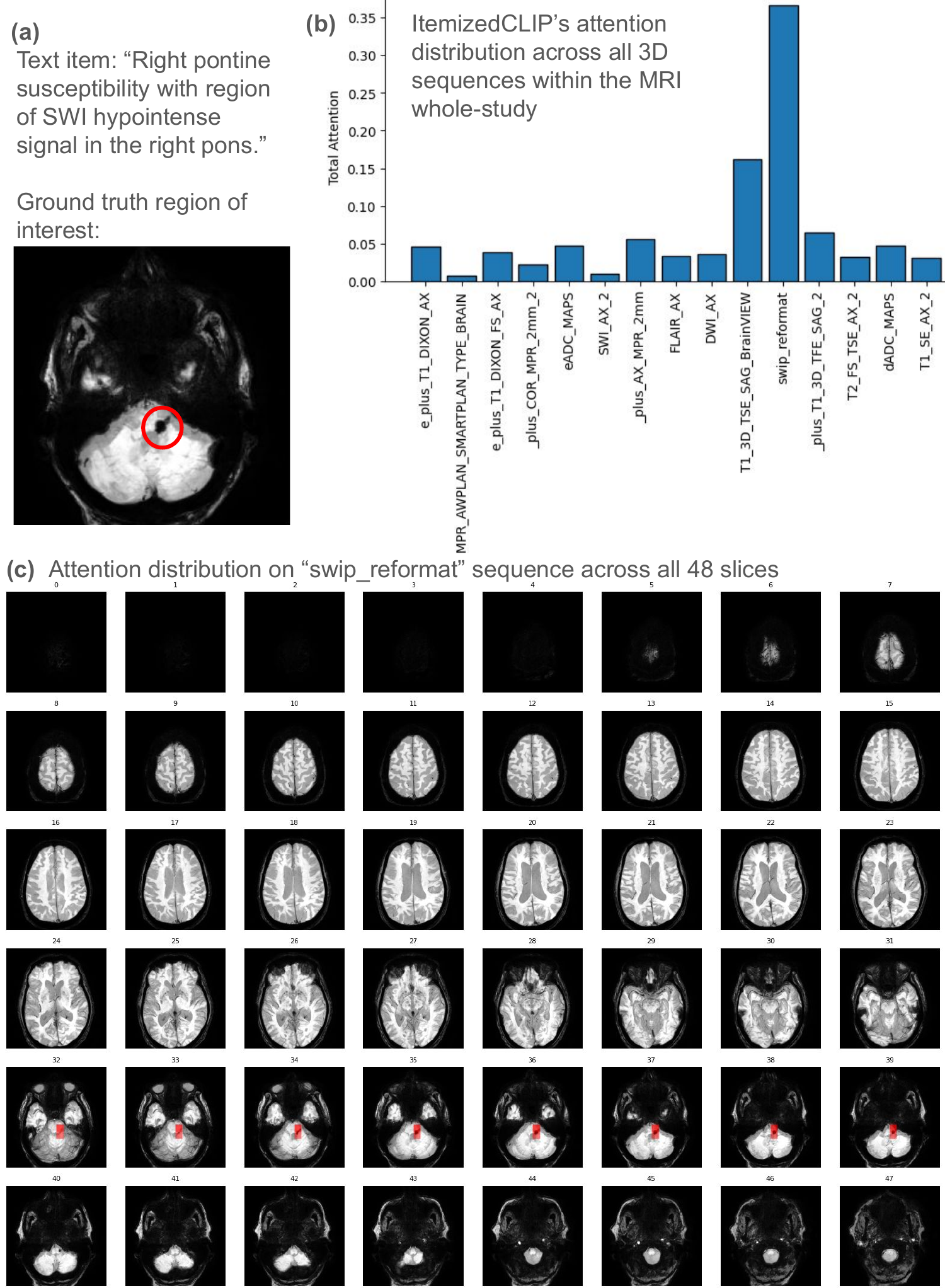}
    \caption{Visualization of attention over an entire brain MRI whole-study. \textbf{(a)} The text item and ground truth region of interest annotation about the text item in the study. \textbf{(b)} Attention distribution across all 3D sequences in the study. \name~correctly focused its attention on an SWI sequence (``swip\_reformat"). \textbf{(c)} Attention distribution across all slices in ``swip\_reformat" sequence. \name~correctly focused its attention over the 2 local visual tokens (each with dimensions 8x16x16) that cover the pontine susceptibility.}
    \label{fig:additionalvis2}
\end{figure*}

We show additional text item attention visualization examples in Figure~\ref{fig:additionalvis1}. In addition, we demonstrate attention visualization over all slices across and within 3D sequences of a brain MRI whole-study in Figure~\ref{fig:additionalvis2}. These examples show that, when given a text item (or any text correctly describing parts of the visuals), \name~can often accurately attend to the correct region of interest associated with the text item, thus the attention visualization makes \name~naturally interpretable and makes its zero-shot decisions more trustworthy.

\subsection{Region-based text retrieval}
\label{app:regionretrieval}

\begin{figure*}
    \centering
    \includegraphics[width=0.95\linewidth]{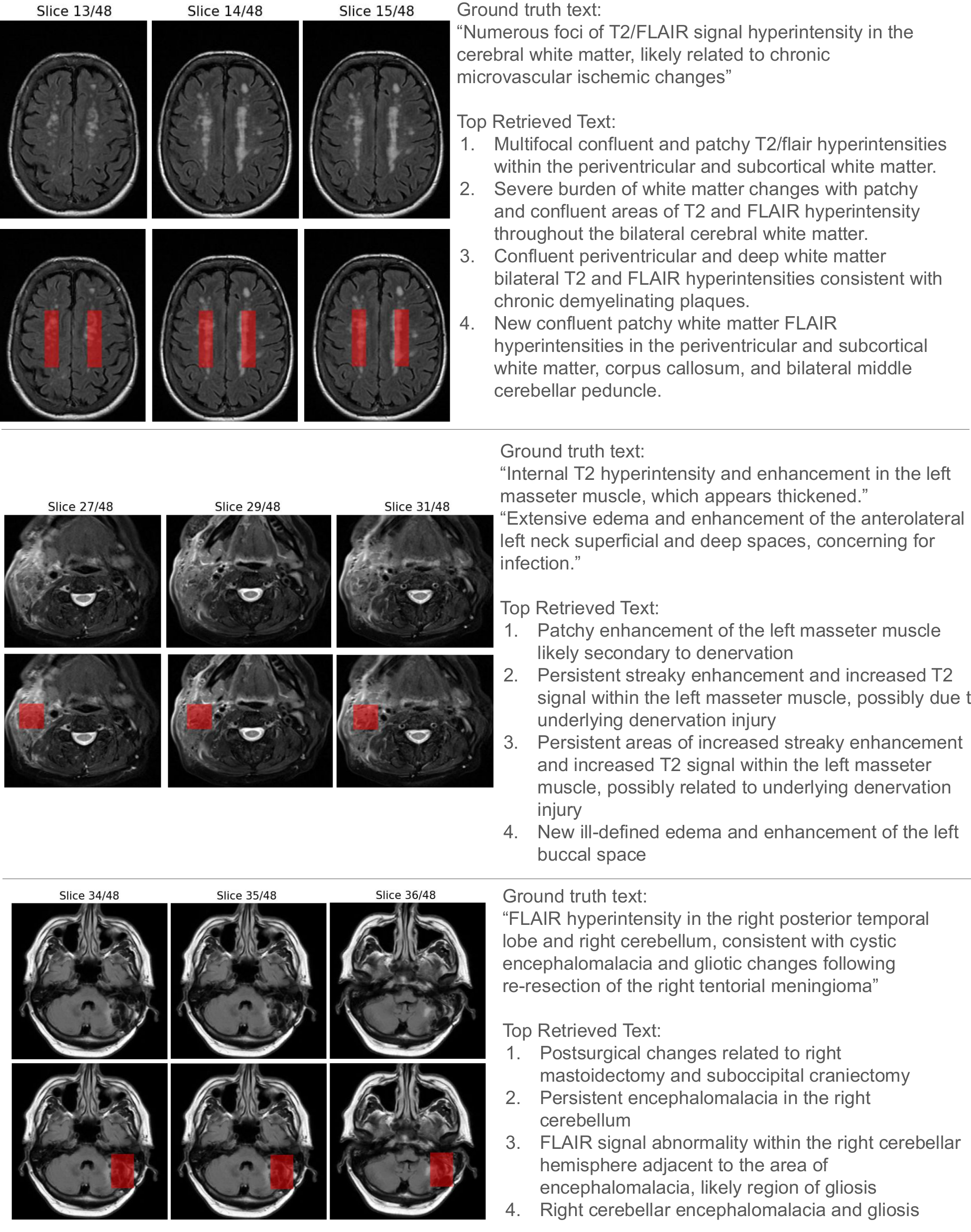}
    \caption{Additional region-based text retrieval examples. We manually select a region in an MRI sequence (highlighted in red), then we use \name~to retrieve text items from all text items in prospective test set, around 100K items in total that best describe this region. We show the ground truth text item from the corresponding radiology report and top 4 retrieved items.}
    \label{fig:regionadditional}
\end{figure*}

We show additional region-based text retrieval examples in Figure~\ref{fig:regionadditional}. For each brain MRI study, we manually select a sequence and a certain region $vp'$ within the sequence (defined in local tokens of size 8x16x16), and we retrieve text items from all text items within the prospective test set ($T$, $\sim$100K items in total) by $\text{argmin}_{t\in T}TCSim(t,vp')$. The retrieved text items often matches well with one or more ground truth text items matched to the study, indicating \name's local visual representation captures both anatomy/location information as well as pathological information. 

\subsection{Item differentiability}
\label{app:differentiability}

\begin{figure*}
    \centering
    \includegraphics[width=0.7\linewidth]{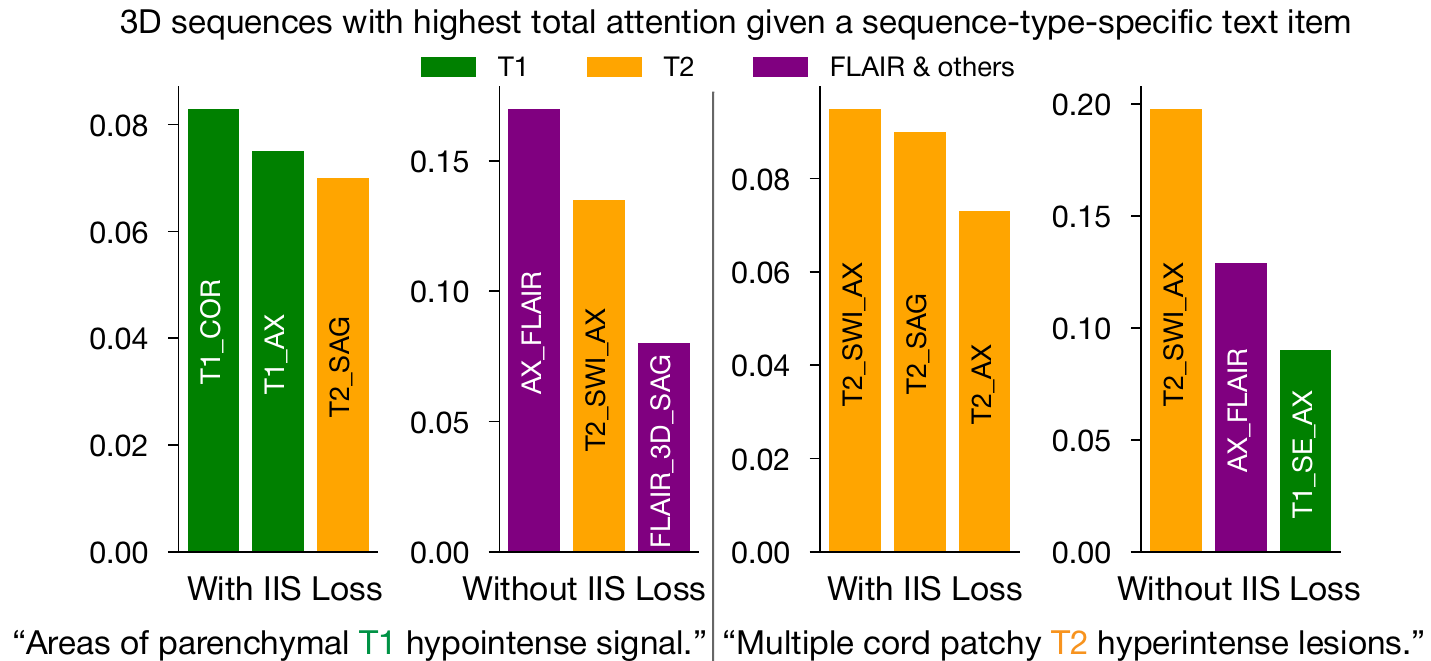}
    \caption{A brain MRI example that demonstrates item differentiability: we show the top-3 3D sequences with the highest total attention (out of 35 3D sequences in the MRI study) over 2 positive text items different text items with 2 different models, one trained with IIS and one without. The model with IIS is able to focus its top attention to the correct sequence types (T1/T2) that each text item refers to, while the model without IIS focus on the same 2 FLAIR/SWI sequences regardless of which type is mentioned in the text item. Please note that FLAIR here refers to a brain MRI sequence type rather than the visual representation learning method~\cite{xiao2025flair}. This example further illustrates that IIS improves item differentiability.}
    \label{fig:differentiabilityapp}
\end{figure*}

We show an additional qualitative example of IIS improving model item differentiability on brain MRI. Each brain MRI whole-study contains many 3D sequences, and sometimes a text item will refer specifically to certain 3D sequence types. In Figure~\ref{fig:differentiabilityapp}, we show an example where two text items for a single study refers to two different types of sequences, T1 and T2. We show the top 3 sequences (out of 35 total sequences in the study) with highest total attention for each text item, on models trained with and without IIS. We see that the model trained with IIS correctly focuses its attention on sequences with types mentioned in the text item, while the model trained without IIS always have the FLAIR and T2-SWI sequences as its top 2 focus. This example further illustrates that IIS guides the model towards better and more accurate item differentiability. (Note that FLAIR here refers to a sequence type in brain MRI, not the visual-language pre-training method FLAIR~\cite{xiao2025flair}.)

\begin{figure*}
    \centering
    \includegraphics[width=\linewidth]{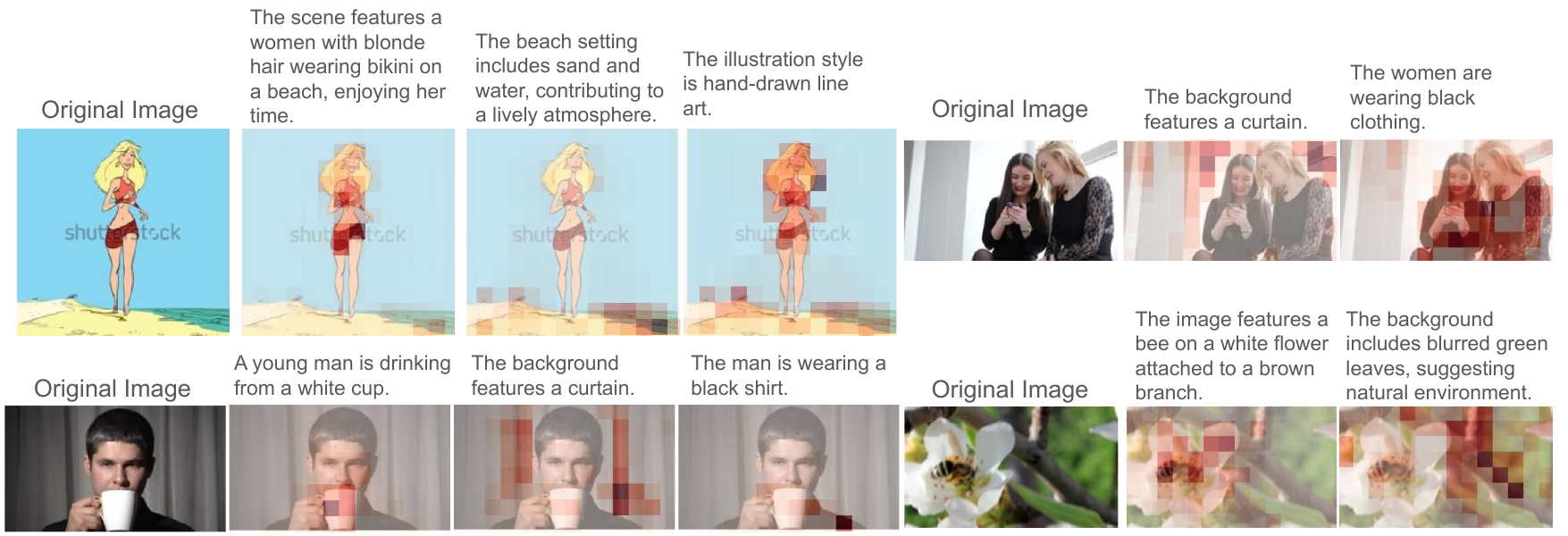}
    \caption{More item differentiability examples on Itemized-cc0.3M. In these examples, \name~is able to accurately visualize different text items by attending to corresponding local visual tokens.}
    \label{fig:additionaldifferentiability}
\end{figure*}

We present more examples on item differentiability of \name~on Itemized-cc0.3M in Figure~\ref{fig:additionaldifferentiability}.

\section{Computational resources}

All experiments in this paper are conducted on a single server with 8 Nvidia L40S GPUs. Training \name~ for whole-study brain MRI on UM220K takes 40 hours. Training \name~for whole-study brain CT on HeadCT240K takes 40 hours. Training \name~for single-sequence Chest CT takes 20 hours. Training \name~for Remote Sensing on RSICD takes 0.5 hours. Training \name~for natural images on Itemized-cc0.3M takes 6 hours. We do not observe significant computation overhead with \name~compared to different combinations of subsets of its components, as all the majority of computation in \name~as well as baselines happens within the visual and language backbones, which are kept the same for fairness of comparison.

%% file: main.bbl
\begin{thebibliography}{42}
\providecommand{\natexlab}[1]{#1}
\providecommand{\url}[1]{\texttt{#1}}
\expandafter\ifx\csname urlstyle\endcsname\relax
  \providecommand{\doi}[1]{doi: #1}\else
  \providecommand{\doi}{doi: \begingroup \urlstyle{rm}\Url}\fi

\bibitem[Baid et~al.(2021)Baid, Ghodasara, Mohan, Bilello, Calabrese, Colak, Farahani, Kalpathy-Cramer, Kitamura, Pati, et~al.]{baid2021brats}
Ujjwal Baid, Satyam Ghodasara, Suyash Mohan, Michel Bilello, Evan Calabrese, Errol Colak, Keyvan Farahani, Jayashree Kalpathy-Cramer, Felipe~C Kitamura, Sarthak Pati, et~al.
\newblock The rsna-asnr-miccai brats 2021 benchmark on brain tumor segmentation and radiogenomic classification.
\newblock \emph{arXiv preprint arXiv:2107.02314}, 2021.

\bibitem[Blankemeier et~al.(2024)Blankemeier, Cohen, Kumar, Van~Veen, Gardezi, Paschali, Chen, Delbrouck, Reis, Truyts, et~al.]{blankemeier2024merlin}
Louis Blankemeier, Joseph~Paul Cohen, Ashwin Kumar, Dave Van~Veen, Syed Jamal~Safdar Gardezi, Magdalini Paschali, Zhihong Chen, Jean-Benoit Delbrouck, Eduardo Reis, Cesar Truyts, et~al.
\newblock Merlin: A vision language foundation model for 3d computed tomography.
\newblock \emph{arXiv preprint arXiv:2406.06512}, 2024.

\bibitem[Boecking et~al.(2022)Boecking, Usuyama, Bannur, Castro, Schwaighofer, Hyland, Wetscherek, Naumann, Nori, Alvarez-Valle, Poon, and Oktay]{boecking2022biomedvlp}
Benedikt Boecking, Naoto Usuyama, Shruthi Bannur, Daniel~C. Castro, Anton Schwaighofer, Stephanie Hyland, Maria Wetscherek, Tristan Naumann, Aditya Nori, Javier Alvarez-Valle, Hoifung Poon, and Ozan Oktay.
\newblock Making the most of text semantics to improve biomedical vision-language processing, 2022.

\bibitem[Cao et~al.(2024)Cao, Zhang, Xia, Mok, Li, Ye, Lu, Zheng, Tang, and Zhang]{cao2024bootstrapping}
Weiwei Cao, Jianpeng Zhang, Yingda Xia, Tony~CW Mok, Zi Li, Xianghua Ye, Le Lu, Jian Zheng, Yuxing Tang, and Ling Zhang.
\newblock Bootstrapping chest ct image understanding by distilling knowledge from x-ray expert models.
\newblock In \emph{Proceedings of the IEEE/CVF Conference on Computer Vision and Pattern Recognition}, pages 11238--11247, 2024.

\bibitem[Chakraborty et~al.(2020)Chakraborty, Bisong, Bhatt, Wagner, Elliott, and Mosconi]{chakraborty-etal-2020-biomedbert}
Souradip Chakraborty, Ekaba Bisong, Shweta Bhatt, Thomas Wagner, Riley Elliott, and Francesco Mosconi.
\newblock {B}io{M}ed{BERT}: A pre-trained biomedical language model for {QA} and {IR}.
\newblock In \emph{Proceedings of the 28th International Conference on Computational Linguistics}, pages 669--679, Barcelona, Spain (Online), 2020. International Committee on Computational Linguistics.

\bibitem[Chilamkurthy et~al.(2018)Chilamkurthy, Ghosh, Tanamala, Biviji, Campeau, Venugopal, Mahajan, Rao, and Warier]{chilamkurthy2018deep}
Sasank Chilamkurthy, Rohit Ghosh, Swetha Tanamala, Mustafa Biviji, Norbert~G Campeau, Vasantha~Kumar Venugopal, Vidur Mahajan, Pooja Rao, and Prashant Warier.
\newblock Deep learning algorithms for detection of critical findings in head ct scans: a retrospective study.
\newblock \emph{The Lancet}, 392\penalty0 (10162):\penalty0 2388--2396, 2018.

\bibitem[Dong et~al.(2023)Dong, Bao, Zheng, Zhang, Chen, Yang, Zeng, Zhang, Yuan, Chen, Wen, and Yu]{dong2023maskclip}
Xiaoyi Dong, Jianmin Bao, Yinglin Zheng, Ting Zhang, Dongdong Chen, Hao Yang, Ming Zeng, Weiming Zhang, Lu Yuan, Dong Chen, Fang Wen, and Nenghai Yu.
\newblock {MaskCLIP}: Masked self-distillation advances contrastive language-image pretraining.
\newblock In \emph{Proceedings of the IEEE/CVF Conference on Computer Vision and Pattern Recognition}, pages 10995--11005, 2023.

\bibitem[Dosovitskiy et~al.(2020)Dosovitskiy, Beyer, Kolesnikov, Weissenborn, Zhai, Unterthiner, Dehghani, Minderer, Heigold, Gelly, Uszkoreit, and Houlsby]{vit}
Alexey Dosovitskiy, Lucas Beyer, Alexander Kolesnikov, Dirk Weissenborn, Xiaohua Zhai, Thomas Unterthiner, Mostafa Dehghani, Matthias Minderer, Georg Heigold, Sylvain Gelly, Jakob Uszkoreit, and Neil Houlsby.
\newblock An image is worth 16x16 words: Transformers for image recognition at scale.
\newblock \emph{CoRR}, abs/2010.11929, 2020.

\bibitem[Draelos et~al.(2021)Draelos, Dov, Mazurowski, Lo, Henao, Rubin, and Carin]{draelos2021machine}
Rachel~Lea Draelos, David Dov, Maciej~A Mazurowski, Joseph~Y Lo, Ricardo Henao, Geoffrey~D Rubin, and Lawrence Carin.
\newblock Machine-learning-based multiple abnormality prediction with large-scale chest computed tomography volumes.
\newblock \emph{Medical image analysis}, 67:\penalty0 101857, 2021.

\bibitem[Fan et~al.(2023)Fan, Krishnan, Isola, Katabi, and Tian]{fan2023improving}
Lijie Fan, Dilip Krishnan, Phillip Isola, Dina Katabi, and Yonglong Tian.
\newblock Improving clip training with language rewrites.
\newblock \emph{Advances in Neural Information Processing Systems}, 36:\penalty0 35544--35575, 2023.

\bibitem[Flanders et~al.(2020)Flanders, Prevedello, Shih, Halabi, Kalpathy-Cramer, Ball, Mongan, Stein, Kitamura, Lungren, et~al.]{flanders2020construction}
Adam~E Flanders, Luciano~M Prevedello, George Shih, Safwan~S Halabi, Jayashree Kalpathy-Cramer, Robyn Ball, John~T Mongan, Anouk Stein, Felipe~C Kitamura, Matthew~P Lungren, et~al.
\newblock Construction of a machine learning dataset through collaboration: the rsna 2019 brain ct hemorrhage challenge.
\newblock \emph{Radiology: Artificial Intelligence}, 2\penalty0 (3):\penalty0 e190211, 2020.

\bibitem[Gao et~al.(2022)Gao, Liu, Xu, Zhang, Li, Ji, and Shen]{gao2022pyramidclip}
Yuting Gao, Jinfeng Liu, Zihan Xu, Jun Zhang, Ke Li, Rongrong Ji, and Chunhua Shen.
\newblock Pyramidclip: Hierarchical feature alignment for vision-language model pretraining.
\newblock \emph{Advances in neural information processing systems}, 35:\penalty0 35959--35970, 2022.

\bibitem[Gao et~al.(2024)Gao, Liu, Xu, Wu, Zhang, Li, Yang, Liu, and Sun]{gao2024softclip}
Yuting Gao, Jinfeng Liu, Zihan Xu, Tong Wu, Enwei Zhang, Ke Li, Jie Yang, Wei Liu, and Xing Sun.
\newblock Softclip: Softer cross-modal alignment makes clip stronger.
\newblock In \emph{Proceedings of the AAAI Conference on Artificial Intelligence}, pages 1860--1868, 2024.

\bibitem[Hamamci et~al.(2024)Hamamci, Er, Almas, Simsek, Esirgun, Dogan, Dasdelen, Durugol, Wittmann, Amiranashvili, et~al.]{hamamci2024developing}
Ibrahim~Ethem Hamamci, Sezgin Er, Furkan Almas, Ayse~Gulnihan Simsek, Sevval~Nil Esirgun, Irem Dogan, Muhammed~Furkan Dasdelen, Omer~Faruk Durugol, Bastian Wittmann, Tamaz Amiranashvili, et~al.
\newblock Developing generalist foundation models from a multimodal dataset for 3d computed tomography.
\newblock \emph{arXiv preprint arXiv:2403.17834}, 2024.

\bibitem[Hammoud et~al.(2024)Hammoud, Itani, Pizzati, Torr, Bibi, and Ghanem]{hammoud2024synthclip}
Hasan Abed Al~Kader Hammoud, Hani Itani, Fabio Pizzati, Philip Torr, Adel Bibi, and Bernard Ghanem.
\newblock Synthclip: Are we ready for a fully synthetic clip training?
\newblock \emph{arXiv preprint arXiv:2402.01832}, 2024.

\bibitem[Ilharco et~al.(2021)Ilharco, Wortsman, Wightman, Gordon, Carlini, Taori, Dave, Shankar, Namkoong, Miller, Hajishirzi, Farhadi, and Schmidt]{ilharco_gabriel_2021_5143773}
Gabriel Ilharco, Mitchell Wortsman, Ross Wightman, Cade Gordon, Nicholas Carlini, Rohan Taori, Achal Dave, Vaishaal Shankar, Hongseok Namkoong, John Miller, Hannaneh Hajishirzi, Ali Farhadi, and Ludwig Schmidt.
\newblock Openclip, 2021.

\bibitem[Jose et~al.(2024)Jose, Moutakanni, Kang, Baldassarre, Darcet, Xu, Li, Szafraniec, Ramamonjisoa, Oquab, Siméoni, Vo, Labatut, and Bojanowski]{jose2024dinov2meetstextunified}
Cijo Jose, Théo Moutakanni, Dahyun Kang, Federico Baldassarre, Timothée Darcet, Hu Xu, Daniel Li, Marc Szafraniec, Michaël Ramamonjisoa, Maxime Oquab, Oriane Siméoni, Huy~V. Vo, Patrick Labatut, and Piotr Bojanowski.
\newblock Dinov2 meets text: A unified framework for image- and pixel-level vision-language alignment, 2024.

\bibitem[Lai et~al.(2024)Lai, Zhang, Zhang, Wu, Bai, Timofeev, Du, Gan, Shan, Chuah, et~al.]{lai2024veclip}
Zhengfeng Lai, Haotian Zhang, Bowen Zhang, Wentao Wu, Haoping Bai, Aleksei Timofeev, Xianzhi Du, Zhe Gan, Jiulong Shan, Chen-Nee Chuah, et~al.
\newblock Veclip: Improving clip training via visual-enriched captions.
\newblock In \emph{European Conference on Computer Vision}, pages 111--127. Springer, 2024.

\bibitem[Lavoie et~al.(2024)Lavoie, Kirichenko, Ibrahim, Assran, Wilson, Courville, and Ballas]{lavoie2024modeling}
Samuel Lavoie, Polina Kirichenko, Mark Ibrahim, Mahmoud Assran, Andrew~Gordon Wilson, Aaron Courville, and Nicolas Ballas.
\newblock Modeling caption diversity in contrastive vision-language pretraining.
\newblock \emph{arXiv preprint arXiv:2405.00740}, 2024.

\bibitem[Lin et~al.(2015)Lin, Maire, Belongie, Bourdev, Girshick, Hays, Perona, Ramanan, Zitnick, and Dollár]{lin2015microsoftcococommonobjects}
Tsung-Yi Lin, Michael Maire, Serge Belongie, Lubomir Bourdev, Ross Girshick, James Hays, Pietro Perona, Deva Ramanan, C.~Lawrence Zitnick, and Piotr Dollár.
\newblock Microsoft coco: Common objects in context, 2015.

\bibitem[Liu et~al.(2024)Liu, Lyu, Lee, and Hollon]{liu2024empirical}
Shixuan Liu, Yiwei Lyu, Honglak Lee, and Todd~C Hollon.
\newblock An empirical study of clip fine-tuning with similarity clusters.
\newblock In \emph{NeurIPS 2024 Workshop on Fine-Tuning in Modern Machine Learning: Principles and Scalability}, 2024.

\bibitem[Liu et~al.(2023)Liu, Wang, Shao, Luo, Qiao, Shou, Zhang, and You]{liu2023mllms}
Yanqing Liu, Kai Wang, Wenqi Shao, Ping Luo, Yu Qiao, Mike~Zheng Shou, Kaipeng Zhang, and Yang You.
\newblock Mllms-augmented visual-language representation learning.
\newblock \emph{arXiv preprint arXiv:2311.18765}, 2023.

\bibitem[Lu et~al.(2017)Lu, Wang, Zheng, and Li]{lu2017exploring}
Xiaoqiang Lu, Binqiang Wang, Xiangtao Zheng, and Xuelong Li.
\newblock Exploring models and data for remote sensing image caption generation.
\newblock \emph{IEEE Transactions on Geoscience and Remote Sensing}, 56\penalty0 (4):\penalty0 2183--2195, 2017.

\bibitem[Lyu et~al.(2026)Lyu, Harake, Chowdury, Banerjee, Gologorsky, Liu, Meissner, Rao, Zhao, Kondepudi, et~al.]{lyu2025learning}
Yiwei Lyu, Samir Harake, Asadur Chowdury, Soumyanil Banerjee, Rachel Gologorsky, Shixuan Liu, Anna-Katharina Meissner, Akshay Rao, Chenhui Zhao, Akhil Kondepudi, et~al.
\newblock Learning neuroimaging models from health system-scale data.
\newblock \emph{Nature Biomedical Engineering}, pages 1--13, 2026.

\bibitem[Mu et~al.(2021)Mu, Kirillov, Wagner, and Xie]{mu2021slip}
Norman Mu, Alexander Kirillov, David Wagner, and Saining Xie.
\newblock Slip: Self-supervision meets language-image pre-training.
\newblock \emph{arXiv preprint arXiv:2112.12750}, 2021.

\bibitem[Nie et~al.(2025)Nie, He, Bie, Wang, Chen, Yang, and Chen]{nie2025conceptclip}
Yuxiang Nie, Sunan He, Yequan Bie, Yihui Wang, Zhixuan Chen, Shu Yang, and Hao Chen.
\newblock Conceptclip: Towards trustworthy medical ai via concept-enhanced contrastive language-image pre-training.
\newblock \emph{arXiv preprint arXiv:2501.15579}, 2025.

\bibitem[Radford et~al.(2021)Radford, Kim, Hallacy, Ramesh, Goh, Agarwal, Sastry, Askell, Mishkin, Clark, et~al.]{radford2021learning}
Alec Radford, Jong~Wook Kim, Chris Hallacy, Aditya Ramesh, Gabriel Goh, Sandhini Agarwal, Girish Sastry, Amanda Askell, Pamela Mishkin, Jack Clark, et~al.
\newblock Learning transferable visual models from natural language supervision.
\newblock In \emph{International conference on machine learning}, pages 8748--8763. PmLR, 2021.

\bibitem[Ribeiro et~al.(2016)Ribeiro, Singh, and Guestrin]{ribeiro2016lime}
Marco~Tulio Ribeiro, Sameer Singh, and Carlos Guestrin.
\newblock "why should {I} trust you?": Explaining the predictions of any classifier.
\newblock In \emph{Proceedings of the 22nd {ACM} {SIGKDD} International Conference on Knowledge Discovery and Data Mining, San Francisco, CA, USA, August 13-17, 2016}, pages 1135--1144, 2016.

\bibitem[R{\"o}sch et~al.(2024)R{\"o}sch, Oswald, Geierhos, and Libovick{\`y}]{rosch2024enhancing}
Philipp~J R{\"o}sch, Norbert Oswald, Michaela Geierhos, and Jind{\v{r}}ich Libovick{\`y}.
\newblock Enhancing conceptual understanding in multimodal contrastive learning through hard negative samples.
\newblock \emph{arXiv preprint arXiv:2403.02875}, 2024.

\bibitem[Sharma et~al.(2018)Sharma, Ding, Goodman, and Soricut]{sharma2018conceptual}
Piyush Sharma, Nan Ding, Sebastian Goodman, and Radu Soricut.
\newblock Conceptual captions: A cleaned, hypernymed, image alt-text dataset for automatic image captioning.
\newblock In \emph{Proceedings of ACL}, 2018.

\bibitem[Shui et~al.(2025)Shui, Zhang, Cao, Wang, Guo, Lu, Yang, Ye, Liang, Zhang, et~al.]{shui2025large}
Zhongyi Shui, Jianpeng Zhang, Weiwei Cao, Sinuo Wang, Ruizhe Guo, Le Lu, Lin Yang, Xianghua Ye, Tingbo Liang, Qi Zhang, et~al.
\newblock Large-scale and fine-grained vision-language pre-training for enhanced ct image understanding.
\newblock \emph{arXiv preprint arXiv:2501.14548}, 2025.

\bibitem[Xiao et~al.(2025)Xiao, Kim, Georgescu, Akata, and Alaniz]{xiao2025flair}
Rui Xiao, Sanghwan Kim, Mariana-Iuliana Georgescu, Zeynep Akata, and Stephan Alaniz.
\newblock Flair: Vlm with fine-grained language-informed image representations.
\newblock In \emph{CVPR}, 2025.

\bibitem[Yang et~al.(2023)Yang, Deng, An, Li, Feng, Guo, Yang, and Liu]{yang2023alip}
Kaicheng Yang, Jiankang Deng, Xiang An, Jiawei Li, Ziyong Feng, Jia Guo, Jing Yang, and Tongliang Liu.
\newblock Alip: Adaptive language-image pre-training with synthetic caption.
\newblock In \emph{Proceedings of the IEEE/CVF International Conference on Computer Vision}, pages 2922--2931, 2023.

\bibitem[Yang et~al.(2024)Yang, Xu, Sellergren, Kohlberger, Zhou, Ktena, Kiraly, Ahmed, Hormozdiari, Jaroensri, et~al.]{yang2024advancing}
Lin Yang, Shawn Xu, Andrew Sellergren, Timo Kohlberger, Yuchen Zhou, Ira Ktena, Atilla Kiraly, Faruk Ahmed, Farhad Hormozdiari, Tiam Jaroensri, et~al.
\newblock Advancing multimodal medical capabilities of gemini.
\newblock \emph{arXiv preprint arXiv:2405.03162}, 2024.

\bibitem[Young et~al.(2014)Young, Lai, Hodosh, and Hockenmaier]{young2014image}
Peter Young, Alice Lai, Micah Hodosh, and Julia Hockenmaier.
\newblock From image descriptions to visual denotations: New similarity metrics for semantic inference over event descriptions.
\newblock \emph{Transactions of the Association for Computational Linguistics}, 2:\penalty0 67--78, 2014.

\bibitem[Yu et~al.(2022)Yu, Wang, Vasudevan, Yeung, Seyedhosseini, and Wu]{yu2022coca}
Jiahui Yu, Zirui Wang, Vijay Vasudevan, Legg Yeung, Mojtaba Seyedhosseini, and Yonghui Wu.
\newblock Coca: Contrastive captioners are image-text foundation models.
\newblock \emph{arXiv preprint arXiv:2205.01917}, 2022.

\bibitem[Yuksekgonul et~al.(2023)Yuksekgonul, Bianchi, Kalluri, Jurafsky, and Zou]{yuksekgonul2023when}
Mert Yuksekgonul, Federico Bianchi, Pratyusha Kalluri, Dan Jurafsky, and James Zou.
\newblock When and why vision-language models behave like bags-of-words, and what to do about it?
\newblock In \emph{International Conference on Learning Representations}, 2023.

\bibitem[Zhai et~al.(2023)Zhai, Mustafa, Kolesnikov, and Beyer]{zhai2023sigmoid}
Xiaohua Zhai, Basil Mustafa, Alexander Kolesnikov, and Lucas Beyer.
\newblock Sigmoid loss for language image pre-training.
\newblock In \emph{Proceedings of the IEEE/CVF international conference on computer vision}, pages 11975--11986, 2023.

\bibitem[Zhang et~al.(2023)Zhang, Xu, Usuyama, Xu, Bagga, Tinn, Preston, Rao, Wei, Valluri, et~al.]{zhang2023biomedclip}
Sheng Zhang, Yanbo Xu, Naoto Usuyama, Hanwen Xu, Jaspreet Bagga, Robert Tinn, Sam Preston, Rajesh Rao, Mu Wei, Naveen Valluri, et~al.
\newblock Biomedclip: a multimodal biomedical foundation model pretrained from fifteen million scientific image-text pairs.
\newblock \emph{arXiv preprint arXiv:2303.00915}, 2023.

\bibitem[Zhao et~al.(2026)Zhao, Lyu, Chowdury, Harake, Kondepudi, Rao, Hou, Lee, and Hollon]{zhao2025scalable}
Chenhui Zhao, Yiwei Lyu, Asadur~Zaman Chowdury, Edward~S Harake, Akhil Kondepudi, Akshay~T Rao, Xinhai Hou, Honglak Lee, and Todd~C Hollon.
\newblock Towards scalable language-image pre-training for 3d medical imaging.
\newblock \emph{Transactions on Machine Learning Research}, 2026.

\bibitem[Zheng et~al.(2024)Zheng, Zhang, Wu, Lu, Ma, Jin, Chen, and Shen]{zheng2024dreamlip}
Kecheng Zheng, Yifei Zhang, Wei Wu, Fan Lu, Shuailei Ma, Xin Jin, Wei Chen, and Yujun Shen.
\newblock Dreamlip: Language-image pre-training with long captions.
\newblock In \emph{European Conference on Computer Vision (ECCV)}. Springer, 2024.

\bibitem[Zhu et~al.(2025)Zhu, Huang, Tang, Musthyala, Yu, Chen, Vega, O'Donnell, Dehkharghani, Frontera, et~al.]{zhu20253d}
Weicheng Zhu, Haoxu Huang, Huanze Tang, Rushabh Musthyala, Boyang Yu, Long Chen, Emilio Vega, Thomas O'Donnell, Seena Dehkharghani, Jennifer~A Frontera, et~al.
\newblock 3d foundation ai model for generalizable disease detection in head computed tomography.
\newblock \emph{arXiv preprint arXiv:2502.02779}, 2025.

\end{thebibliography}
